\documentclass{article}
\PassOptionsToPackage{table}{xcolor}
\usepackage{microtype}
\usepackage{graphicx}
\usepackage{subcaption}
\usepackage{booktabs} 

\usepackage{hyperref}

\usepackage[accepted]{icml2026}
\makeatletter
\let\ICML@appearing\relax
\makeatother

\usepackage{amssymb}
\usepackage{mathtools}
\usepackage{amsthm}

\usepackage[capitalize,noabbrev]{cleveref}

\usepackage{xcolor}
\usepackage{amsfonts}  
\usepackage{amsmath}

\usepackage{booktabs}
\usepackage[most]{tcolorbox}
\usepackage{placeins}

\usepackage{tabularx}
\usepackage{ragged2e}
\definecolor{promptbg}{RGB}{242,246,250}   
\definecolor{promptframe}{RGB}{165,175,185}
\definecolor{prompttitle}{RGB}{30,30,30}    

\theoremstyle{plain}

\theoremstyle{definition}

\theoremstyle{remark}

\usepackage[textsize=tiny]{todonotes}

\icmltitlerunning{Game-Theoretic Co-Evolution for LLM-Based Heuristic Discovery}

\begin{document}

\twocolumn[
  \icmltitle{Game-Theoretic Co-Evolution for LLM-Based Heuristic Discovery}



  \icmlsetsymbol{equal}{*}

  \begin{icmlauthorlist}
    \icmlauthor{Xinyi Ke}{casia}
    \icmlauthor{Kai Li}{casia,ucas}
    \icmlauthor{Junliang Xing}{tsinghua}
    \icmlauthor{Yifan Zhang}{casia,ucas}
    \icmlauthor{Jian Cheng}{casia,AiRiA}

  \end{icmlauthorlist}

  \icmlaffiliation{casia}{C\textsuperscript{2}DL, Institute of Automation, Chinese Academy of Sciences}
  \icmlaffiliation{ucas}{School of Artificial Intelligence, University of Chinese Academy of Sciences}
  \icmlaffiliation{tsinghua}{Tsinghua University}
  \icmlaffiliation{AiRiA}{AiRiA}

  \icmlcorrespondingauthor{Kai Li}{kai.li@ia.ac.cn}

  \icmlkeywords{Machine Learning, ICML}

  \vskip 0.3in
]



\printAffiliationsAndNotice{}  

\begin{abstract}

Large language models (LLMs) have enabled rapid progress in automatic heuristic discovery (AHD), yet most existing methods are predominantly limited by static evaluation against fixed instance distributions, leading to potential overfitting and poor generalization under distributional shifts.
We propose Algorithm Space Response Oracles (ASRO), a game-theoretic framework that reframes heuristic discovery as a program level co-evolution between solver and instance generator. ASRO models their interaction as a two-player zero-sum game, maintains growing strategy pools on both sides, and iteratively expands them via LLM-based best-response oracles against mixed opponent meta-strategies, thereby replacing static evaluation with an adaptive, self-generated curriculum. Across multiple combinatorial optimization domains, ASRO consistently outperforms static-training AHD baselines built on the same program search mechanisms, achieving substantially improved generalization and robustness on diverse and out-of-distribution instances.
\end{abstract}

\section{Introduction}
Combinatorial Optimization (CO) underpins critical applications across engineering, autonomous systems, and large-scale resource management, yet its NP-hard nature makes exact algorithms infeasible at realistic scales~\cite{papadimitriou1998combinatorial}. Modern solvers therefore rely on hand-crafted heuristics and metaheuristics~\cite{blum_metaheuristics_2003, burke_exploring_2009,burke_classification_2010, 
drake_recent_2020}. This manual design process is slow, brittle, and biased toward the structural assumptions of human experts, which often leads to severe performance degradation under distributional or structural shift~\cite{bengio_machine_2021, manchanda_generalization_2023}.

The emergence of Large Language Models (LLMs) capable of synthesizing executable programs~\cite{austin_program_2021} has opened a promising direction for Automatic Heuristic Design (AHD). 
Recent work leverages this capability to automate the synthesis of executable heuristic solvers,
enabling rapid construction of competitive solvers and partially automating the manual algorithm-design pipeline. 

In practice, existing LLM-AHD pipelines adopt a closed-loop generate–evaluate–refine workflow: the LLM proposes heuristic programs, their execution on a fixed evaluator yields performance and behavioral feedback, and this feedback guides subsequent refinement. This iterative loop forms the backbone of current LLM–AHD methodologies
~\cite{romera-paredes_mathematical_2024,liu_evolution_2024,ye_reevo_2024,novikov_alphaevolve_2025}.
Despite their empirical success, these frameworks are fundamentally tied to static evaluation~\cite{liu_evolution_2024}: heuristics are optimized against a fixed evaluation distribution, e.g., pre-designed datasets.

This static setting raises two core limitations.
(i) It invites overfitting and fragility: discovered heuristics optimized on a fixed evaluator often generalize poorly under distributional shift~\cite{sim_hype_2025}.
(ii) It induces a performance ceiling: once a heuristic is near-optimal, the evaluator fails to expose new weaknesses, hindering further improvement.
Even when moving beyond static evaluation, existing approaches often rely on hand-crafted curricula or ad hoc difficulty schedules, leaving instance adaptation outside the core optimization objective and structurally retaining a single-agent training paradigm.
Consequently, solver–instance interactions lack a principled, unified optimization framework, making stable and sustained co-adaptation difficult.

Motivated by this, we propose and formalize a game-theoretic, co-evolutionary formulation of LLM-based heuristic discovery, termed the Algorithm Space Response Oracle (ASRO).
In ASRO, solvers and instance generators are modeled as two players in a zero-sum game, with strategies represented as executable programs\footnote{Throughout the paper, we study algorithms through their executable program realizations. For notational simplicity, we refer to these realizations as \emph{programs}, which serve as concrete, executable solvers or generators synthesized by the LLM.}. 
Solvers map instances to solutions, while generators induce instance distributions that challenge solver performance.
ASRO maintains growing pools of strategies on both sides, and iteratively (i) evaluates all solver–generator pairs to construct a payoff matrix, 
(ii) computes mixed meta-strategies (i.e., probability distributions over the current strategy pools), 
and (iii) invokes LLM-based best-response oracles to synthesize new solver or generator programs against the opponent’s meta-strategy.

Conceptually, ASRO draws inspiration from Policy Space Response Oracles (PSRO)~\cite{lanctot_unified_2017}, sharing its population-based, best-response-driven optimization paradigm.
Unlike PSRO, which operates in parametric policy spaces and approximates best responses via reinforcement learning, ASRO works directly in discrete program space, where strategies are executable solver and instance generator programs synthesized by LLMs.
A key advantage of the programmatic formulation is that  generators are executable code, where a single program induces a structured stochastic family of instance distributions.
By embedding such generators into the meta-game, ASRO couples instance difficulty and diversity within a unified co-evolutionary process, where generators shift their induced distributions as solver mixtures improve, yielding solver pools that are robust beyond specific benchmarks.
ASRO induces a self-generated curriculum: generators adaptively expose weaknesses of the current solver mixture, and solvers in turn adapt to increasingly hard and diverse instance distributions.

Our main contributions are as follows:
\begin{itemize}
    \item \textbf{Game-theoretic framework for LLM-based AHD.} We formulate LLM-driven heuristic discovery as a two-player zero-sum game over executable programs, replacing static evaluation with a PSRO-style ``solve-and-expand'' loop in discrete program space.
    \item \textbf{Meta-game driven co-evolution.} Persistent strategy pools and equilibrium-based meta-strategies induce sustained solver--generator co-evolution, avoiding short-horizon adversarial dynamics.  
    \item \textbf{An oracle-agnostic, program-space framework.} ASRO specifies a game-theoretic optimization structure in program space while leaving the choice of program search mechanism open.
    \item \textbf{Empirical validation on multiple CO domains.} Across three representative CO domains, ASRO consistently outperforms static-training counterparts built on identical program search mechanisms, achieving improved robustness and generalization under distributional and structural shifts.
\end{itemize}

\section{Related Works}
\subsection{Automated Heuristic Design}
Early AHD relied on hyper-heuristics and genetic programming, evolving heuristics expressed in manually designed domain-specific languages (DSLs)~\cite{burke_hyperheuristics_2003, burke_hyperheuristics_2013,burke_classification_2019,pillay_hyperheuristics_2018}.
A major line of work develops heuristic expressions via Genetic Programming (GP), spanning early program-evolution frameworks \cite{koza_genetic_1992,burke_exploring_2009, burke_genetic_2010} 
and more recent applications to routing and packing problems~\cite{hildebrandt_improved_2010, duflo_gp_2019}.
These methods are often paired with automatic configuration frameworks such as SMAC and \text{irace}~\cite{hutter_sequential_2011,lopez-ibanez_irace_2016a}, but their
low-level search spaces limit scalability and prevent the
exploitation of high-level algorithmic structures.

The integration of LLMs into AHD has transformed the field, enabling the synthesis of executable heuristics through semantic reasoning. 
A prominent paradigm is evaluation-guided, iterative program evolution, exemplified by a range of representative works, including early frameworks such as AEL~\cite{liu_algorithm_2023} and FunSearch~\cite{romera-paredes_mathematical_2024}, as well as EoH~\cite{liu_evolution_2024} and more recent systems like AlphaEvolve~\cite{novikov_alphaevolve_2025}.
Meanwhile, works like~\cite{ye_reevo_2024,liu_experienceguided_2025} further enrich this process by incorporating verbal reflection and historical experience to guide improvement, reflecting a broader trend toward iterative self-refinement~\cite{madaan_selfrefine_2023,shinn_reflexion_2023}.
Expanding the scope further, research has investigated the automatic generation and evolution of metaheuristic algorithms and operators using LLMs across various optimization paradigms~\cite{pluhacek_leveraging_2023, stein_llamea_2025,hemberg_evolving_2024,dat_hsevo_2025, yao_multiobjective_2025, surina_algorithm_2025}.
However, these approaches remain predominantly evaluator-static, optimizing against predefined task distributions and overlooking the potential of co-evolutionary adaptation.

\subsection{Generalization and Robustness in Combinatorial Optimization}
Generalization has long been a central focus in Neural Combinatorial Optimization (NCO). Prior work explores curriculum learning \cite{lisicki_evaluating_2020, zhang_learning_2022,iklassov_study_2023,liu_cl4co_2024}, distributional diversification \cite{jiang_learning_2022,zhou_omnigeneralizable_2023,luo_neural_2023}, and hardness-aware or adversarial sampling \cite{zhang_learning_2022}, including game-theoretic adversarial variants \cite{wang_gametheoretic_2022, wang_asp_2024}. Furthermore, meta-learning 
\cite{manchanda_generalization_2023,qiu_dimes_2022a,wang_unsupervised_2022,son_metasage_2023,chen_efficient_2023}
and knowledge distillation \cite{bi_learning_2022, zhang_neural_2023, zheng_mtlkd_2025} have been widely adopted to enhance robustness. 
Notably, these lines of work primarily operate by optimizing continuous parameters within neural architectures, rather than evolving symbolic, executable programs.

Recent LLM-AHD studies examine generalization via heuristic set diversification, cross task transfer, and problem agnostic modeling \cite{jiang_llmopt_2024,anoy_generalizable_2025,liu_eohs_2025, chen_improving_2025}. 
Most, however, rely on static distributions. While EALG introduces an initial GAN-style solver–generator loop \cite{duan_ealg_2025}, its self-play formulation does not explicitly maintain a persistent strategy set on either side.
This makes the dynamics more prone to cycling~\cite{czarnecki2020real} and limits the ability to form a stable, equilibrium-driven curriculum across diverse instance families, which is precisely the gap that our ASRO framework aims to address.

\subsection{Game-Theoretic Response Frameworks}
Game-theoretic response frameworks comprise a family of methods that iteratively expand a game’s strategy sets by solving restricted games and introducing best responses. Classical approaches like Fictitious Play, Double Oracle, and empirical game-theoretic analysis established this foundation~\cite{brown_iterative_1951,mcmahan_planning_2003,wellman_methods_2006,shoham_multiagent_2008}, while PSRO later instantiated these principles in deep multi-agent reinforcement learning by maintaining and updating sets of policies via iterative best-response training~\cite{lanctot_unified_2017}. 
This line of work was subsequently refined by improved meta-solvers and ranking mechanisms—exemplified by Alpha-PSRO~\cite{muller_generalized_2019} and Nash averaging \cite{balduzzi_mechanics_2018,balduzzi_openended_2019}—and scaled to complex environments via AlphaStar’s league-based training \cite{jaderberg_population_2017, vinyals_grandmaster_2019}.
In contrast to these approaches, which operate on parametric neural policies, our work adapts the response-oracle paradigm to the domain of program synthesis, where solver logic and instance generators co-evolve as executable programs.

\section{ASRO: A Program Space Co-Evolution Framework for AHD}
\FloatBarrier
\begin{figure*}[t!]
    \centering
    \includegraphics[width=0.85\textwidth]{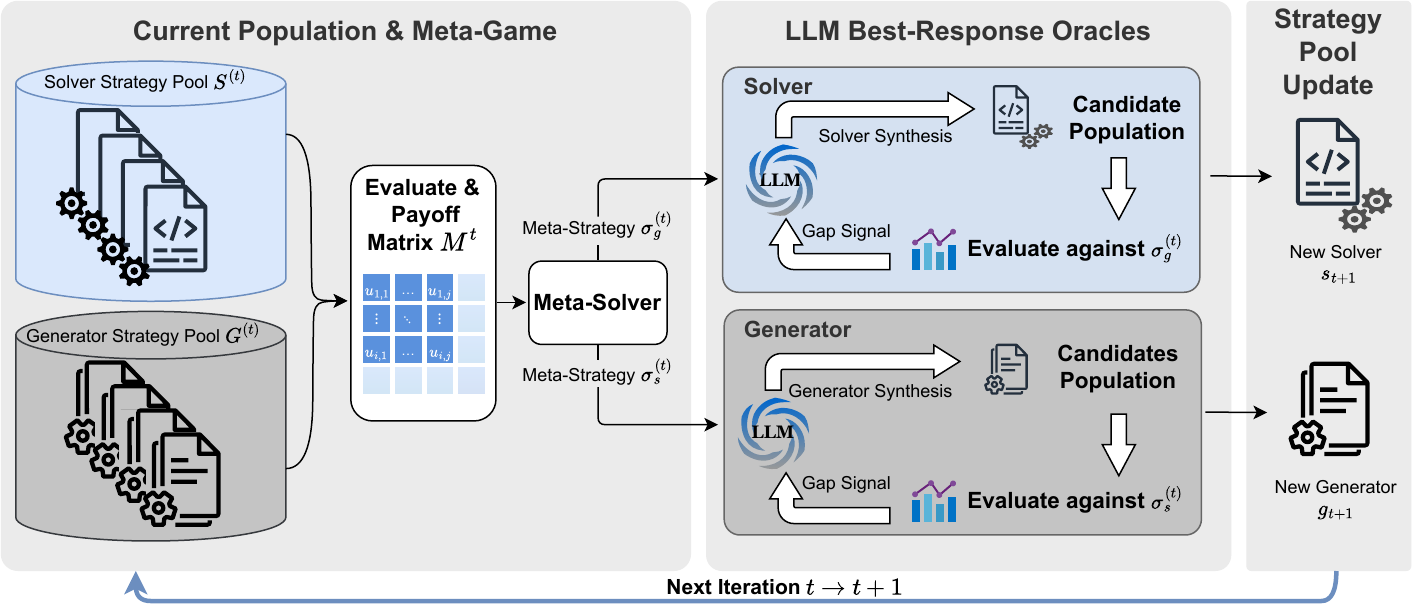}
     \caption{Overview of ASRO.
    The iterative process is structured into three phases.
    (Left) The current solver strategy pool $\mathcal{S}^{(t)}$ and generator strategy pool $\mathcal{G}^{(t)}$ are evaluated pairwise to compute a payoff matrix $M^{(t)}$.
    A meta-solver then computes mixed meta-strategies $(\sigma_s^{(t)}, \sigma_g^{(t)})$ over the existing strategy pools.
    (Middle) Program-space best-response oracles synthesize new solver and generator programs conditioned on the opponent’s current meta-strategy, instantiated by an underlying program search mechanism (see Figure~\ref{fig:eoh_oracle}).
    (Right) The newly discovered solver $s_{t+1}$ and generator $g_{t+1}$ are added to their respective strategy pools for the next iteration ($t \to t+1$), enabling systematic exploration of the program space.}
    \label{fig:asro_overview}
\end{figure*}

\subsection{Preliminaries: Policy-Space Response Oracles}
PSRO \citep{lanctot_unified_2017} is a response-based framework for $N$-player games that operates by iteratively expanding a restricted set of strategies $\Pi_i^{(t)} \subset \Pi_i$ for each player $i = 1,\ldots,N$. 
The procedure follows an iterative ``solve-and-expand'' workflow.
At iteration $t$, a \textbf{meta-strategy} $\sigma_i^{(t)}$ is computed by solving
the game restricted to the current strategy sets, where $\sigma_i^{(t)}$ is a
probability distribution over $\Pi_i^{(t)}$ and $\sigma_{-i}^{(t)}$ denotes the
joint meta-strategy of all players except player $i$.
Subsequently, a Best-Response (BR) oracle expands the strategy set by finding
\begin{equation} \label{eq:psro_br}
\pi_{i, \mathrm{new}} = \arg\max_{\pi_i \in \Pi_i}
\mathbb{E}_{\pi_{-i} \sim \sigma_{-i}^{(t)}} \left[ u_i(\pi_i, \pi_{-i}) \right],
\end{equation}
where $u_i(\pi_i, \pi_{-i})$ is the expected utility of player $i$.
In PSRO, strategies are typically parameterized by neural policies, with BRs approximated via reinforcement learning.
Details are provided in Appendix~\ref{app:psro}.

\subsection{A Program-Space Game Formulation of AHD}
Our key observation is that AHD implicitly defines an interaction between two programmatic choices: a solver program that determines how solutions are constructed for a CO instance, and an instance-generation program that determines which instances are used for evaluation. Based on this observation, we introduce the ASRO framework, which extends PSRO from 
policy spaces to the space of symbolic, executable programs.

\paragraph{Strategy Spaces.}
The \textbf{Solver Space} $\mathcal{S}$ contains solver programs $s$, each of which specifies heuristic decision logic that is executed within a fixed, domain-specific solver procedure to construct a feasible solution for a given instance $x$.
The \textbf{Generator Space} $\mathcal{G}$ contains generator programs $g$ that specify stochastic instance distributions; sampling $x \sim g(\cdot)$ yields structurally valid problem instances generated from the program’s specification.

\paragraph{Payoff Function.}
We model the interaction between solver and generator programs as a two-player zero-sum game defined over a task-specific \emph{normalized reference gap}, a standard performance measure in CO~\cite{beasley_orlibrary_1990}.
For a solver program $s$ and an instance $x$, let
\begin{equation}
\mathrm{gap}(s,x)
=
\frac{V(s,x) - v^*(x)}{v^*(x)}
\label{eq:gap}
\end{equation}
denote the normalized gap with respect to a reference value $v^*(x)$, where $V(s,x)$ is the objective value produced by solver $s$ on instance $x$, and $v^*(x)$ denotes a task-dependent reference value.
Smaller gaps indicate better solver performance.
The payoff of a solver--generator pair $(s,g)$ is defined as the expected gap under the generator-induced instance distribution:
\begin{equation}
\label{eq:payoff}
U(s,g)
\;=\;
\mathbb{E}_{x \sim g}\big[\, \mathrm{gap}(s,x) \,\big].
\end{equation}
The solver seeks to minimize $U(s,g)$, while the generator seeks to maximize it.

\subsection{Meta-Strategy Computation}
Given the current solver and generator strategy pools $\mathcal{S}^{(t)}$ and $\mathcal{G}^{(t)}$, ASRO constructs a restricted payoff matrix
$M^{(t)}$, where each entry is
\begin{equation}
    M^{(t)}_{ij} = U(s_i, g_j).
\end{equation}

Since solvers minimize gap while generators maximize it, the interaction forms a finite zero-sum matrix game.
At iteration $t$, ASRO computes mixed meta-strategies $(\sigma_s^{(t)}, \sigma_g^{(t)})$ by approximately solving the minimax equilibrium
\begin{equation}
    \min_{\sigma_s} \max_{\sigma_g}\;
    \mathbb{E}_{s \sim \sigma_s,\, g \sim \sigma_g}
    \big[\, M^{(t)}_{sg} \,\big].
\end{equation} 
For finite zero-sum games, this equilibrium can be obtained via the standard linear-program formulation of the minimax problem or, 
equivalently, via no-regret dynamics such as multiplicative weights~\cite{freund_decisiontheoretic_1997}.
The resulting meta-strategies provide opponent distributions for the subsequent best-response synthesis.

\subsection{Program-Space Best-Response Oracle}
\label{subsec:br_oracle}
To optimize the best-response objective in Eq.~(\ref{eq:psro_br}) over executable programs, ASRO assumes access to a \emph{program-space best-response oracle},
implemented by a \emph{program search mechanism}.
Given the opponent’s meta-strategy, this oracle approximately optimizes the induced best-response objective and returns a corresponding program, referred to as an \emph{approximate best response} (ABR).
When the responding player is a solver or a generator, we write the returned programs as 
$\textsc{ABR}_s(\sigma_g)$ and $\textsc{ABR}_g(\sigma_s)$, respectively.
ASRO is agnostic to how this search is performed, enabling modular instantiations.
We denote an instantiation using mechanism $X$ as \emph{ASRO-X}.
$X$ may correspond to a mechanism derived from or inspired by existing LLM-based algorithm discovery methods, without inheriting their original optimization objectives or training pipelines.

\begin{figure}[t!]
    \centering
    \includegraphics[width=0.9\linewidth]{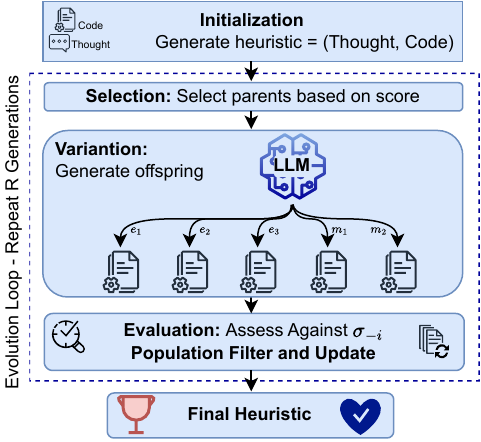}
    \caption{
    Program-space best-response oracle in ASRO, instantiated via an EoH-style evolutionary program search.
    Under a fixed opponent meta-strategy $\sigma_{-i}$ provided by ASRO, the oracle performs a bounded evolutionary search over programs in the corresponding strategy space.
    Candidate programs are initialized, explored, and mutated using structured LLM-based operators, and evaluated according to the induced best-response objective.
    }
    \label{fig:eoh_oracle}
\end{figure}

\textbf{Instantiation with Evolutionary Search (ASRO-EoH).}
In this work, ASRO instantiates the best-response oracle via a bounded evolutionary program search derived from Evolution of Heuristics (EoH)~\cite{liu_evolution_2024}, referred to as \emph{ASRO-EoH}; an overview is shown in Figure~\ref{fig:eoh_oracle}.
The oracle maintains a population of $K$ heuristic programs, which is initialized at the first ASRO iteration and warm-started from previously discovered programs thereafter.
Within a single oracle call, the population is evolved for a fixed number of $R$ rounds using structured exploration and mutation operators to generate offspring programs,  followed by a selection step over the current population and newly generated variants that balances performance under the best-response objective and population diversity.
After the search terminates, the highest-performing program is returned as an approximate best response and used to expand the ASRO strategy pool.
Implementation details are provided in Appendix~\ref{app:EoH}.

\subsection{Training Procedure and Analysis}
ASRO (Figure~\ref{fig:asro_overview}) is an iterative framework that alternates between solving a restricted program-level meta-game and expanding the strategy pools via LLM-based approximate best responses. 
At each iteration, ASRO updates the payoff matrix over the current solver and generator strategy pools, computes equilibrium meta-strategies, 
and invokes best-response oracles to synthesize new solver and generator programs.
The full procedure is summarized in Algorithm~\ref{alg:asro}.

\begin{algorithm}[t!]
\caption{ASRO: Program-Level Co-Evolution for AHD}
\label{alg:asro}
\begin{algorithmic}[1]
\STATE \textbf{Input:} Initial pools $\mathcal{S}^{(0)}$, $\mathcal{G}^{(0)}$; iteration horizon $T$
\FOR{$t = 0$ to $T-1$}
    \STATE \textcolor{gray}{(1) Meta-Game Evaluation}
    \STATE Update payoff matrix $M^{(t)}$ over the strategy pools

    \STATE \textcolor{gray}{(2) Meta-Strategy Computation}
    \STATE $(\sigma_s^{(t)}, \sigma_g^{(t)}) \gets \textsc{MetaSolver}(M^{(t)})$

    \STATE \textcolor{gray}{(3) LLM-Based Approximate Best Responses}
    \STATE $s_{\mathrm{new}} \gets \mathrm{ABR}_s(\sigma_g^{(t)})$ \hfill (solver BR)
    \STATE $g_{\mathrm{new}} \gets \mathrm{ABR}_g(\sigma_s^{(t)})$ \hfill (generator BR)
    
    \STATE \textcolor{gray}{(4) Strategy Pool Expansion}
    \STATE $\mathcal{S}^{(t+1)} \gets \mathcal{S}^{(t)} \cup \{ s_{\mathrm{new}} \}$
    \STATE $\mathcal{G}^{(t+1)} \gets \mathcal{G}^{(t)} \cup \{ g_{\mathrm{new}} \}$
\ENDFOR
\STATE \textbf{Output:} Final solver and generator strategy pools
\end{algorithmic}
\end{algorithm}

\section{Experiments}
\label{sec:experiments}
\subsection{Experimental Setup}
\textbf{Common Settings.}
We use \textbf{DeepSeek-V3.2}~\cite{deepseek-ai_deepseekv32_2025} as the backbone LLM for both solver and generator synthesis, with a temperature of $1$, promoting exploration in program space. The co-evolutionary process runs for $T=8$ iterations. 
During training, a small portion of instances is always drawn from a fixed, domain-specific \emph{ base generator} to stabilize early-stage co-evolution and avoid premature over-specialization

\textbf{Framework-level comparison with EoH.}
ASRO-EoH and EoH use the same evolutionary program search mechanism; the difference is that EoH is trained on a fixed distribution induced by the base generator, whereas ASRO-EoH operates under a game-based framework with an evolving instance distribution induced by the generator strategy pool.

\textbf{Evaluation Metric.}
Across all tasks, we evaluate performance using the reference gap defined in Eq.~(\ref{eq:gap}).

\textbf{Convergence Metrics.}
We assess convergence using exploitability-based metrics estimated using approximate best responses, which quantify the gain from a unilateral deviation.
Given mixed strategies $(\sigma_s^{(t)}, \sigma_g^{(t)})$ at iteration $t$, the solver and generator exploitabilities are
\begin{equation}
\begin{aligned}
\mathrm{Exp}_s(t)
&=
u(\sigma_s^{(t)}, \sigma_g^{(t)})
-
u(\mathrm{ABR}_s(\sigma_g^{(t)}), \sigma_g^{(t)}), \\
\mathrm{Exp}_g(t)
&=
u(\sigma_s^{(t)}, \mathrm{ABR}_g(\sigma_s^{(t)}))
-
u(\sigma_s^{(t)}, \sigma_g^{(t)}),
\end{aligned}
\end{equation}
We report the Approximate NashConv (ANC),
$\mathrm{ANC}(t)=\mathrm{Exp}_s(t)+\mathrm{Exp}_g(t)$,
which summarizes the exploitability of the current solver--generator mixture~\cite{timbers_approximate_2022}.

\subsection{Domains and Implementation Details}
We evaluate ASRO on three well-studied combinatorial optimization problems: online bin packing (OBP), the traveling salesman problem (TSP), and the capacitated vehicle routing problem (CVRP), spanning online and offline optimization as well as diverse decision structures.
Hyperparameter settings and LLM usage are summarized in Appendix~\ref{app:details}, while task-specific formulations, prompt templates, and domain details are provided in Appendix~\ref{app:OBP}, \ref{app:TSP}, and~\ref{app:CVRP}.

\subsubsection{Online Bin Packing (OBP)}
\textbf{Problem Setting.}
We study online bin packing problem \cite{garey_approximation_1981, johnson_worstcase_1974}, where items arrive sequentially and must be assigned irrevocably to bins of capacity $C$, with the objective of minimizing the total number of bins used \cite{balogh_optimal_2015, seiden_online_2002}.

\textbf{Solver Representation.}
Solver programs are executed within a fixed online packing framework. 
Each solver program specifies an online greedy decision rule in the form of a numerical priority function that assigns a score to each feasible bin upon item arrival~\cite{ramanan_online_1989}.
The packing process sequentially handles arriving items and enforces feasibility; 
all other components of the procedure are kept fixed, with items placed into the feasible bin with the highest score, without lookahead or post-processing.

\textbf{Reference Values.}
$v^*(x)$ are obtained using OR-Tools CP-SAT when optimality can be certified~\cite{googleor-toolsteam_python_2024}; otherwise, we use the Martello--Toth lower bound~\cite{martello_lower_1990}.

\textbf{Benchmarks and Arrival Orders.}
We evaluate on widely used public bin packing benchmarks~\cite{delorme_bpplib_2018}: Falkenauer \texttt{T} and \texttt{U} families~\cite{falkenauer_hybrid_1996}, the Hard28 benchmark~\cite{delorme_bpplib_2018}, and additional instances drawn from a range of Weibull distributions~\cite{castineiras_weibullbased_2012}. 
These benchmarks cover both classical test cases and structurally challenging settings.
The base generator used during training follows a Weibull distribution.
We evaluate under multiple standard arrival orders (random, ascending), fixed independently of the algorithms.

\textbf{Baselines.}
We compare ASRO against canonical online heuristics (First Fit~\cite{Ullman1971performance, dosa_first_2013} and Best Fit~\cite{johnson_worstcase_1974, albers_best_2020}) and EoH~\cite{liu_evolution_2024}, an LLM-based solver-only baseline.

\subsubsection{Traveling Salesman Problem (TSP)}
\textbf{Problem Setting.}
We consider the Euclidean Traveling Salesman Problem with $N$ cities, where the objective is to find a Hamiltonian tour of minimum total length.

\textbf{Solver Representation.}
Solver programs operate within a fixed Guided Local Search (GLS) framework~\cite{voudouris_guided_1999a, liu_experienceguided_2025}.
Each solver specifies an edge-distance update rule, producing a modified distance matrix for evaluating candidate moves.
All other components of GLS, including the search loop and standard local search operators (e.g., 2-opt, relocate), are kept fixed, while the learned update reshapes the search landscape.

\textbf{Reference Values.}
$v^*(x)$ are obtained using the Concorde solver for all instances considered~\cite{applegate_traveling_2006}.

\textbf{Benchmarks.}
Evaluation is conducted on held-out uniformly sampled 100-node instances and TSPLIB~\cite{reinelt_tsplib_1991}, covering homogeneous Euclidean cases and structured instances with diverse geometric properties. The base generator samples instances uniformly.

\textbf{Baselines.}
We compare ASRO against classical constructive heuristics (Nearest and Farthest Insertion; NI, FI)~\cite{rosenkrantz_analysis_2009} and EoH~\cite{liu_evolution_2024}.

\subsubsection{Capacitated Vehicle Routing Problem}
\textbf{Problem Setting.}
We study the CVRP, where a fleet of identical vehicles departs from and returns to a single depot to serve customers with known demands.
Each customer must be visited exactly once, and the total demand served by any vehicle cannot exceed the vehicle capacity $Q$.
The objective is to minimize the total travel cost over all routes.

\textbf{Solver Representation.}
Solver programs are executed within a fixed greedy route-construction procedure.
Each solver program specifies a numerical priority function that assigns a score to each currently feasible customer at a routing step.
The customer with the highest score is selected and appended to the current route.

\textbf{Reference Values.}
Reference values $v^*(x)$ are obtained using PyVRP~\cite{wouda_pyvrp_2024} under a fixed time budget, serving as high-quality heuristic reference solutions and not necessarily global optima.

\textbf{Benchmarks.}
We evaluate on CVRP benchmark instances from CVRPLIB \cite{augerat_computational_1995, uchoa_new_2017}, including the A, B, E, F, M, P, and X families. 
These instance families span a broad range of problem sizes and structural characteristics, covering both relatively simple and highly challenging routing scenarios. 
The base generator samples customer coordinates uniformly from $[0,100]^2$ and customer demands uniformly from $[1, Q/3]$.

\textbf{Baselines.}
We compare ASRO with several classical CVRP heuristics, including nearest-neighbor construction with local improvement (NN+2-opt) and parallel insertion heuristics (PI)~\cite{liu_heuristics_2023}, as well as EoH~\cite{liu_evolution_2024}.

\subsection{Results}
\begin{figure}[t]
    \centering
    \includegraphics[width=1\linewidth]{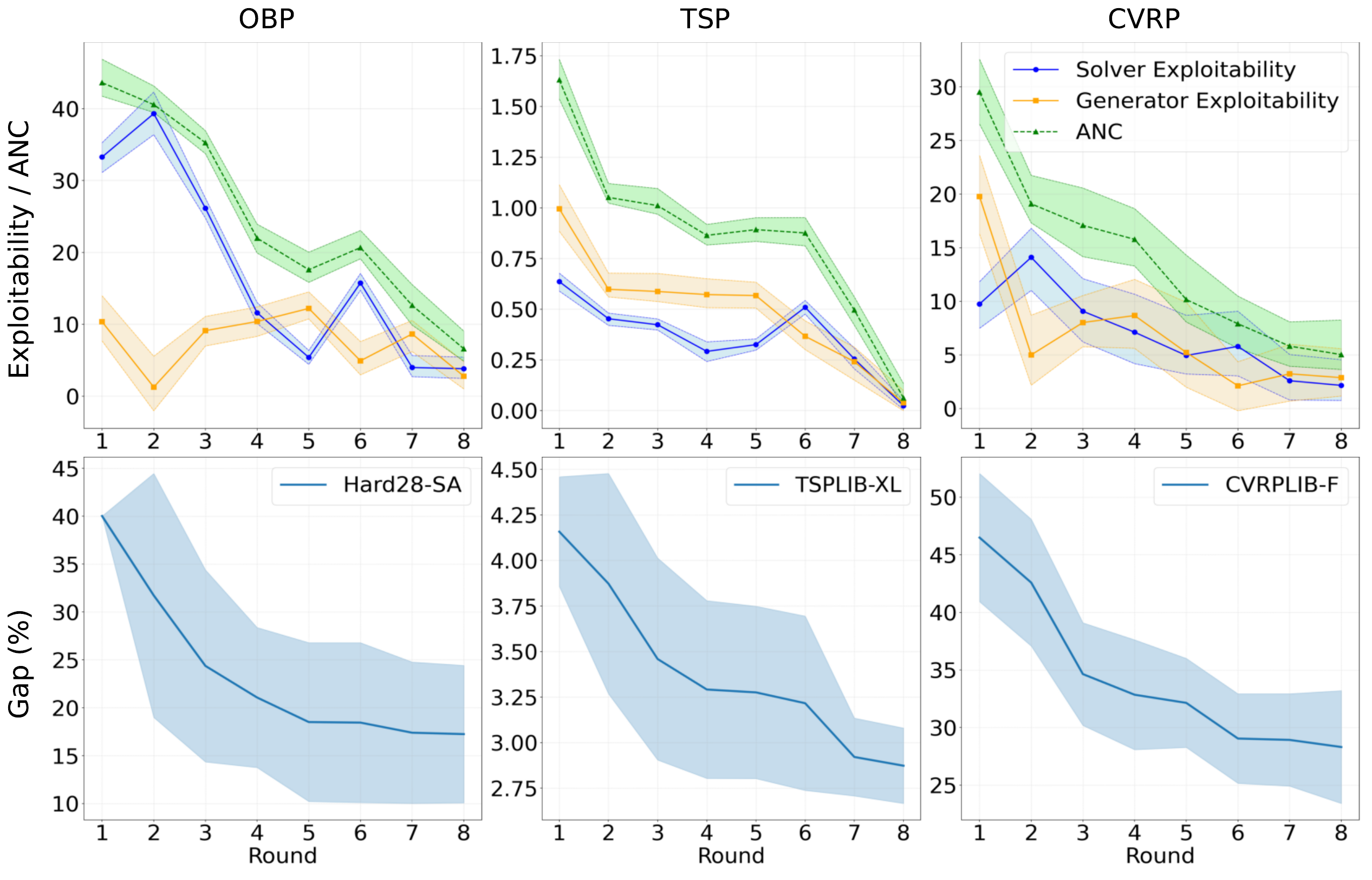}
    \caption{\textbf{ASRO training dynamics across combinatorial optimization tasks.}
    (\textbf{Top}) 
    Convergence behavior of ASRO on OBP, TSP, and CVRP, measured by solver exploitability, generator exploitability, and ANC over iterations, shown for a representative ASRO run; shaded regions denote confidence intervals via bootstrap resampling over instances (20 per generator).
    (\textbf{Bottom}) Stability across random runs: 
    test gap (\%) (mean $\pm$ std) over five independent runs on the hardest benchmark for each task, where benchmark difficulty is determined by the EoH baseline ranking.
    }
    \label{fig:asro_dynamics}
\end{figure}

Figure~\ref{fig:asro_dynamics} summarizes the empirical training dynamics and stability of ASRO across different combinatorial optimization tasks.
Across domains, ANC exhibits a general downward trend over iterations, while solver and generator exploitability remain bounded.
Meanwhile, performance, measured by the optimality gap on test benchmarks, improves consistently across random runs despite stochastic LLM synthesis and approximate best-response updates.
Task-specific differences are evident: OBP exhibits stronger local fluctuations due to its discrete and non-smooth payoff structure; TSP shows comparatively smoother dynamics consistent with its geometric formulation and local search procedures; and CVRP admits greater room for solver–generator adaptation, reflecting its intrinsic structure induces a non-trivial trade-off between minimizing route cost and maintaining capacity feasibility, which gives rise to more persistent adaptive interactions without destabilizing overall convergence.

\begin{figure}[t]
    \centering
    \includegraphics[width=1\linewidth]{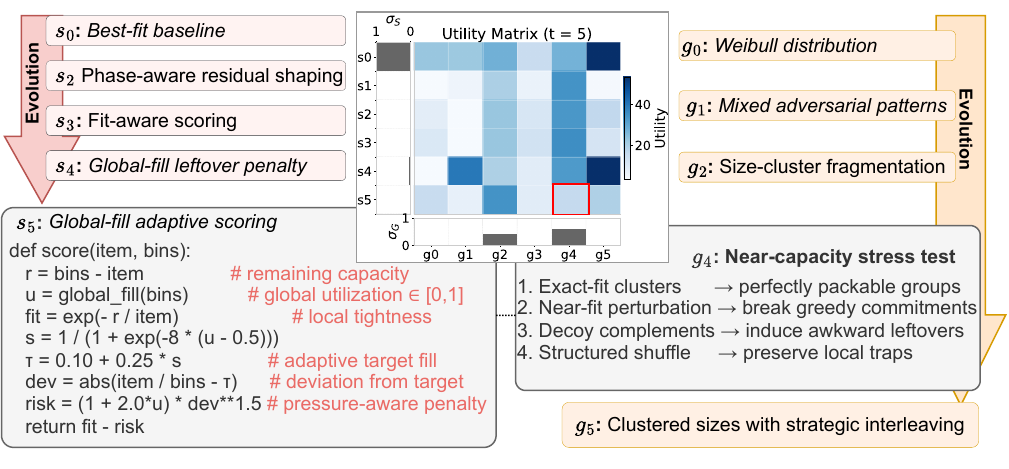 }   
    \caption{\textbf{A snapshot of solver--generator co-evolution in online bin packing.}
    The figure shows a utility matrix at Round~5, with meta-strategies computed from Round~4.
    Side panels summarize the solver and generator strategies present in the pool at Round~4.
    In this snapshot, the generator-side meta-strategy assigns dominant weight to $g_4$, and the solver $s_5$ is generated as a best response under this adversarial pressure.
    The dashed connection indicates that $s_5$ is generated as a best response to $g_4$.
    The highlighted entry (red box) shows the utility of $s_5$ evaluated against $g_4$.
    }   
    \label{fig:obp_metagame}
\end{figure}

\begin{table}[t]
\centering
\caption{\textbf{Main OBP results (gap \% $\downarrow$).}
Gap to the lower bound on standard OBP benchmarks.
F-T and F-U denote the Falkenauer random-arrival sets.
Hard28-R and Hard28-SA correspond to random and size-ascending arrivals on the Hard28 benchmark.
}
\small
\setlength{\tabcolsep}{3pt} 
\renewcommand{\arraystretch}{1.15}

\begin{tabular}{@{\hspace{1mm}}lccccc@{\hspace{1mm}}}
\toprule
Method & F-U & F-T & Hard28-R & Hard28-SA & Weibull \\
\midrule
Best Fit  & 5.39 & 12.66  & 8.45 & 40.02 & 1.75 \\
First Fit  & 6.21 & 12.63 & 9.88 & 40.02 & 2.02 \\
\midrule
EoH & 5.00 & 12.11 & 8.40 & 40.02 & 1.56 \\
\midrule
\rowcolor{lightgray!40}
ASRO-EoH & \textbf{4.53} & \textbf{7.11}  & \textbf{8.06} & \textbf{12.45} & \textbf{1.20} \\

\bottomrule
\end{tabular}%
\label{tab:obp_main}
\end{table}

\paragraph{OBP}
Table~\ref{tab:obp_main} summarizes the performance on standard online bin packing benchmarks.
Across all settings, ASRO consistently outperforms the EoH baseline.
As the instance structure becomes more challenging—from the simpler Falkenauer U set, to the more structured Falkenauer T set, and further to the Hard28 benchmark—the performance gap between ASRO and EoH tends to increase.
This trend suggests that adversarial co-evolution yields greater benefits on harder and more structured instances.
Importantly, even on the Weibull distribution used to train EoH,
ASRO does not exhibit performance degradation, suggesting improved robustness without sacrificing in-distribution performance.

Figure~\ref{fig:obp_metagame} provides a view of how ASRO induces solver adaptation through generator pressure.
The generator meta-strategy concentrates on $g_4$, which constructs item sequences that appear easy to pack early on but systematically result in fragmented bin capacity at later stages.
In this setting, such residual-trap structures are particularly harmful to locally greedy solvers, whose irrevocable early decisions leave little flexibility for future items.
Facing this concentrated stress pattern, the solver-side oracle synthesizes $s_5$ as an approximate best response.
Compared to earlier heuristics, $s_5$ appears to trade off immediate fit against future leftover risk, favoring packing decisions that preserve global flexibility under near-capacity conditions.
This interaction provides an illustrative example of the core mechanism of ASRO: generators expose concrete failure modes of the solver mixture, and solvers adapt by internalizing defenses against these modes in program space.

\paragraph{TSP}
Table~\ref{tab:tsp_main_table} shows that ASRO consistently outperforms the EoH baseline on both simple uniform instances and more complex TSPLIB benchmarks.
Moreover, as problem size increases, the performance gap between ASRO and EoH widens, indicating stronger scalability and improved generalization to larger instances, even though both methods are trained with generators limited to the same problem size.
On several instances, including PR107, PR136, PR144, TSP225, U159, and KROB100, 
ASRO attains small negative optimality gaps relative to reported best-known solution (BKS) values for TSPLIB instances~\cite{reinelt_tsplib_1991}, suggesting highly competitive solution quality under the same evaluation protocol.
Overall, these results demonstrate that ASRO achieves robust and scalable performance across TSP instances of varying structural complexity and problem size.

\begin{table}[t]
\centering
\caption{\textbf{Main TSP results (gap \% $\downarrow$).}
Average optimality gap on a synthetic uniform distribution and on TSPLIB instances grouped by size: 
LIB-S ($n \le 200$), LIB-M ($201 \le n \le 500$), LIB-L ($501 \le n \le 1000$), and LIB-XL ($n > 1000$).}

\small
\setlength{\tabcolsep}{5pt} 
\renewcommand{\arraystretch}{1.15}

\begin{tabular}{@{\hspace{1mm}}lrrrrr@{\hspace{1mm}}}
\toprule
Method
& LIB-S
& LIB-M
& LIB-L
& LIB-XL
& Uniform \\
\midrule
NI
& 19.84
& 23.51
& 23.03
& 21.22
& 20.88 \\
FI
& 7.30
& 10.49
& 13.30
& 15.19
& 8.54 \\
\midrule
EoH
& 0.77
& 1.99
& 3.40
& 4.20
& 0.27 \\
\midrule
\rowcolor{lightgray!40}
ASRO-EoH
& \textbf{0.21}
& \textbf{0.49}
& \textbf{1.43}
& \textbf{3.00}
& \textbf{0.05} \\
\bottomrule
\end{tabular}%
\label{tab:tsp_main_table}
\end{table}

\begin{table}[t]
\centering
\caption{\textbf{Main CVRP results (gap \% $\downarrow$).}
Average optimality gap across CVRPLib instance families (A, B, E, F, M, P, and X).
}

\small
\setlength{\tabcolsep}{3pt} 
\renewcommand{\arraystretch}{1.15}

\begin{tabular}{@{\hspace{1mm}}lccccccc@{\hspace{1mm}}}
\toprule
Method & A & B & E & F & M & P & X \\
\midrule
NN + 2-opt
& 35.58 & 41.30 & 34.81 & 46.53 & 47.14 & 28.24 & 22.71 \\
PI
& 41.82 & 39.39 & 63.07 & 43.73 & 95.15 & 51.48 & 58.34 \\
\midrule
EoH
& 20.19 & 12.22 & 41.83 & 59.78 & 54.22 & 23.68 & 25.71 \\
\midrule
\rowcolor{lightgray!40}
ASRO-EoH
& \textbf{16.85} & \textbf{10.49} & \textbf{20.21} & \textbf{28.71}
& \textbf{22.45} & \textbf{19.45} & \textbf{13.65} \\
\bottomrule
\end{tabular}%
\label{tab:cvrp_main}
\end{table}

\paragraph{CVRP}
Table~\ref{tab:cvrp_main} reports average optimality gaps on CVRPLib instance families.
While A and B consist of relatively regular and well-structured instances,
E, F, M, P, and X exhibit substantially higher heterogeneity in customer distributions, route lengths, and capacity profiles. 
ASRO achieves the lowest gap across all families, performing strongly on both simpler and more complex instances, indicating robust performance across diverse CVRP instance structures.

\subsection{Ablation Study}
\label{sec:ablation}

\paragraph{Data Augmentation and Self-play.}
To examine whether the gains of ASRO can be attributed to stronger training data or to adversarial interaction alone, we compare ASRO against two EoH-based baselines: 
(i) EoH with data augmentation (DA) and (ii) an EoH self-play variant.
In both cases, the program search mechanism is identical to ASRO’s, ensuring that differences arise from instance distributions or interaction dynamics rather than from search.

For DA, the solver is trained on a fixed mixture of geometric instance families, including uniform, clustered, grid-like, ring-shaped, and multi-scale distributions, providing substantially richer distributional diversity than single-distribution training. Full sampling details are provided in Appendix~\ref{app:tsp_da}.
The self-play baseline removes the game-theoretic structure of ASRO while retaining solver--generator optimization. At each iteration, the solver and generator are optimized against each other, but no historical strategies are retained. In ASRO terminology, this corresponds to discarding the strategy pool and restricting both players to their most recent strategies, resulting in a greedy, memoryless adversarial loop.

As shown in Appendix Table~\ref{tab:tsp_ablation_da}, both data augmentation and self-play yield clear improvements over standard EoH, indicating that increased distributional diversity and direct adversarial interaction are beneficial for heuristic learning. However, both variants remain inferior to ASRO, especially on larger and more structured TSPLIB instances. This suggests that while these factors contribute to stronger solvers, they are insufficient on their own: the persistent strategy pool and meta-game dynamics in ASRO play a key role in maintaining sustained pressure and robust co-evolution.

\textbf{Effect of LLM Backbones.}
Appendix Table~\ref{tab:tsp_ablation_llm} reports ASRO-EoH with different LLM backbones for program search.
While stronger backbones achieve modestly lower gaps on TSPLIB instances, smaller models remain effective. Notably, ASRO-EoH with LLaMA-3-8B~\cite{grattafiori_llama_2024} outperforms the EoH baseline trained with a stronger backbone (DeepSeek) on most evaluation settings, indicating that ASRO’s gains primarily stem from solver–generator co-evolution rather than backbone scale.

\textbf{Oracle-agnosticity of ASRO.}
All main experiments instantiate the program-space best-response oracle using an EoH-style program search mechanism; however, ASRO does not depend on any particular choice of search procedure. 
Replacing this instantiation with a ReEvo-style mechanism~\cite{ye_reevo_2024} consistently improves over the corresponding ReEvo baseline on TSPLIB (Appendix Table~\ref{tab:tsplib_reevo}), indicating that ASRO’s gains arise from its response-based evaluation structure rather than from any particular search mechanism.

\textbf{Computation Cost Analysis.}
ASRO incurs additional computational cost in best-response search, since each iteration requires LLM-driven program-space exploration 
under opponent mixed strategies and repeated evaluations of candidate programs.
However, this cost is largely amortized by parallel and batched evaluation, as detailed in Appendix~\ref{app:batch_eval}.
To better characterize this overhead, we report an equal-time comparison between ASRO and EoH in Appendix Table~\ref{tab:time_budget}, where both methods are evaluated under the same wall-clock time budget.
Under this constraint, ASRO consistently achieves lower optimality gaps across all regimes.
One key factor is that EoH converges quickly under a fixed instance distribution, resulting in diminishing returns, whereas ASRO maintains additional improvement capacity through alternating solver and generator updates.

\section{Discussion}
\paragraph{Summary.}
This paper proposed ASRO, a game-theoretic framework for automatic heuristic discovery that replaces static evaluation with adaptive, interaction-driven assessment by modeling solver–generator interaction as a two-player zero-sum game with persistent strategy pools.
We evaluated ASRO on three representative combinatorial optimization problems, including online bin packing, the traveling salesman problem, and the capacitated vehicle routing problem.
Experimental results show that ASRO consistently discovers heuristics that outperform the EoH baseline, which is trained on fixed instance distributions using the same LLM-based search procedure, and demonstrates strong robustness across diverse and structured benchmark instances.
In practice, ASRO trades additional computation for improved robustness, while its evaluation and search procedures remain naturally amenable to large-scale parallelization.
Overall, ASRO establishes a principled game-theoretic framework that formulates heuristic discovery as an interactive process over executable programs, enabling robustness and generalization beyond fixed evaluation settings.

\textbf{Limitations and future work.}
ASRO currently relies on reliable performance signals to compare generator strategies, yet for many combinatorial optimization instances exact optima are unavailable; in such cases, optimality gaps must be computed against strong but incomplete or time-limited oracles (e.g., CP-SAT bounds~\cite{googleor-toolsteam_python_2024} or feasible solutions from PyVRP~\cite{wouda_pyvrp_2024}). Such approximations introduce estimation noise into the meta-game and may obscure fine-grained distinctions among generators. Looking forward, ASRO can be extended beyond a strictly zero-sum formulation to multi-objective or regularized meta-games that explicitly trade off hardness, diversity, and realism for more controlled distribution shaping; it can also be extended to support teacher–student interactions, where generators serve as adaptive curriculum designers rather than purely adversarial opponents; and more broadly, the program-space game perspective underlying ASRO may generalize beyond combinatorial optimization to settings such as algorithmic reasoning, planning, and symbolic decision-making.

\section*{Impact Statement}
This paper presents work whose goal is to advance the field of Machine
Learning. There are many potential societal consequences of our work, none
which we feel must be specifically highlighted here.

\bibliography{main}
\bibliographystyle{icml2026}

\newpage
\appendix
\onecolumn
\section{Policy-Space Response Oracle}
\label{app:psro}

\subsection{Game-Theoretic Setup}
We consider a general $N$-player normal-form game. 
Each player $i \in \{1,\dots,N\}$ chooses a strategy $\pi_i$ from a (possibly infinite) strategy space $\Pi_i$. 
The joint strategy profile is denoted by $\pi = (\pi_1,\dots,\pi_N)$, and $\pi_{-i}$ refers to the strategies of all players except $i$. 
Player $i$ receives utility $u_i(\pi_i, \pi_{-i}) \in \mathbb{R}$.

\subsection{Restricted Games and Meta-Strategies}
Directly solving games over the full strategy spaces $\Pi_i$ is generally intractable. 
PSRO~\cite{lanctot_unified_2017} addresses this by iteratively constructing a \emph{restricted game}.
At iteration $t$, each player $i$ maintains a finite set of candidate strategies $\Pi_i^{(t)} \subset \Pi_i$.
The restricted game is defined by evaluating all joint strategy profiles drawn from
$\Pi_1^{(t)} \times \cdots \times \Pi_N^{(t)}$. 
A \emph{meta-strategy} $\sigma_i^{(t)}$ is a mixed strategy over $\Pi_i^{(t)}$, i.e., a probability distribution satisfying
\[
\sigma_i^{(t)} \in \Delta(\Pi_i^{(t)}).
\]
Here, $\Delta(\Pi_i^{(t)})$ denotes the probability simplex over the finite set $\Pi_i^{(t)}$.
In practical implementations, a meta-strategy often samples a strategy $\pi_i \sim \sigma_i^{(t)}$ and uses it for evaluation within that iteration.

When the restricted game is finite, computing meta-strategies reduces to solving a standard equilibrium problem over a payoff matrix. 
For clarity, we illustrate meta-strategy computation in a two-player zero-sum setting.
In this case, the game is fully characterized by the utility of player~1, with $u_2(\pi_1, \pi_2) = -u_1(\pi_1, \pi_2)$.
The restricted game can therefore be represented by a payoff matrix
$M^{(t)} \in \mathbb{R}^{m \times n}$ with entries
\[
M^{(t)}_{ij} = u_1(\pi_1^i, \pi_2^j),
\]
where $\pi_1^i \in \Pi_1^{(t)}$ and $\pi_2^j \in \Pi_2^{(t)}$.

The meta-strategy for player~1 can then be obtained by solving the following linear program,
as implied by the minimax theorem for two-player zero-sum games~\cite{neumann_theory_1959, shoham_multiagent_2008}: 

\begin{align*}
\max_{x,\, v} \quad & v \\
\text{s.t.} \quad
& x^\top M^{(t)} \ge v \mathbf{1}, \\
& x \ge 0, \quad \mathbf{1}^\top x = 1,
\end{align*} 

where $x$ corresponds to the mixed strategy $\sigma_1^{(t)}$ over $\Pi_1^{(t)}$. 
An analogous linear program yields the meta-strategy for player~2.

\subsection{Best-Response Oracles}
Given opponents’ meta-strategies $\sigma_{-i}^{(t)}$, a best-response (BR) for player $i$ is defined as a strategy that maximizes expected utility against the induced distribution over opponent strategies:
\[
\pi_{i,\mathrm{BR}} \in
\arg\max_{\pi \in \Pi_i}
\mathbb{E}_{\pi_{-i} \sim \sigma_{-i}^{(t)}}
\left[ u_i(\pi, \pi_{-i}) \right].
\]
In practice, computing exact best responses is rarely tractable. PSRO therefore relies on approximate BR oracles~\cite{lanctot_unified_2017}, whose role is to return a high-performing strategy under the current opponent meta-strategy, rather than a provably optimal one.
The above definition assumes utility maximization; minimization objectives can be handled by negating the payoff without loss of generality.

\subsection{PSRO Algorithm}
PSRO proceeds in an iterative ``solve-and-expand'' loop:
\begin{enumerate}
    \item \textbf{Initialization.}
    Initialize each population $\Pi_i^{(0)}$ with a small set of strategies.
    
    \item \textbf{Meta-game solving.}
    At iteration $t$, construct the restricted game induced by
    $\{\Pi_i^{(t)}\}_{i=1}^N$ and compute a meta-strategy
    $\sigma^{(t)} = (\sigma_1^{(t)}, \dots, \sigma_N^{(t)})$
    using a suitable equilibrium solver.
    
    \item \textbf{Response-based expansion.}
    For one or more players, invoke a BR oracle against $\sigma_{-i}^{(t)}$
    to obtain a new strategy, which is added to the population:
    \[
    \Pi_i^{(t+1)} \leftarrow \Pi_i^{(t)} \cup \{\pi_{i,\mathrm{new}}\}.
    \]
\end{enumerate}
By iteratively expanding the restricted game with targeted best responses,
PSRO refines the approximation of the underlying full game.

\section{EoH-style Program Search Mechanism}
\label{app:EoH}
Within ASRO, we instantiate the program-space best-response oracle using an EoH-style evolutionary program search mechanism~\cite{liu_evolution_2024}.
Conditioned on the opponent’s meta-strategy $\sigma^{(t)}_{-i}$, this mechanism evolves a population of solver or generator programs via LLM-driven variation and selection to approximately optimize the induced best-response objective.

In this mechanism, each program is represented through a natural-language “thought” that guides
LLM reasoning and an executable “code” component that is directly evaluated. This dual representation enables both conceptual and syntactic variation during evolution in practice.

\subsection{Evolutionary Operators and Prompt Strategies}
\label{app:eoh_ops}
The EoH-style search mechanism includes one initialization operator and five variation operators, grouped into exploration and modification categories to generate diverse and progressively refined heuristics.

\textbf{Initialization (I1).}
At the first ASRO iteration, I1 constructs the initial population by prompting the LLM to produce heuristics entirely from scratch.  
This provides a starting population for subsequent evolution.

Exploration and modification operators follow a unified usage pattern: in each evolutionary round, a parent set of size $p$ is sampled from the current population; an operator-specific prompt is applied; and the LLM outputs a new heuristic derived from these parents.

\textbf{Exploration Operator E1 (Diverse Exploration).}
E1 encourages large jumps in program space by asking the LLM to design a heuristic that differs substantially from all selected parents.  
Its role is to introduce large, conceptually novel variations that sustain population diversity.

\textbf{Exploration Operator E2 (Shared-Idea Recombination).}
E2 performs semantic recombination.  
The LLM is instructed to identify the common idea shared by the parents and then construct a new heuristic that preserves this idea while introducing new structural elements.  
This operator encourages abstraction and meaningful recombination of successful patterns.

\textbf{Modification Operator M1 (Semantic Refinement).}
M1 improves a single parent by refining its underlying idea.  
The LLM analyzes the heuristic’s logic and proposes targeted adjustments that strengthen its conceptual soundness or operational behavior.

\textbf{Modification Operator M2 (Parameterized Adjustment).}
M2 performs local numerical tuning.  
It directs the LLM to adjust constants, thresholds, or coefficients without altering the heuristic’s overall structure, enabling fine-grained performance improvements.

\textbf{Modification Operator M3 (Simplification and Pruning).}
M3 simplifies a parent heuristic by identifying and removing redundant or overly complex components while preserving the core idea.  
This promotes parsimony and often yields heuristics with stronger generalization.

\begin{figure}[t]
\centering
\fbox{
\begin{minipage}{0.92\linewidth}
\textbf{EoH Prompt Template Structure}

\vspace{4pt}
\textbf{1. Task Description}\\
Defines the optimization problem and the required I/O format. 

\vspace{4pt}
\textbf{2. Operator-Specific Prompt}\\
Instruction for the chosen operator (I1, E1, E2, M1, M2, M3). Specifies how the LLM should reason over parent heuristics and produce a variant.

\vspace{4pt}
\textbf{3. Expected Output}\\
LLM must first describe the heuristic idea, then provide a full Python function with a fixed signature (name, inputs, outputs).

\vspace{4pt}
\textbf{4. Note}\\
Additional constraints: determinism, no extra imports, concise response, and strict adherence to the required code format.

\vspace{4pt}
\textbf{5. Parent Heuristics}\\
Includes the thought and code of $p$ parent heuristics (empty only in I1). Serves as in-context examples for reasoning and variation.
\end{minipage}
}
\caption{Unified structure of all prompt templates used by EoH evolutionary operators. Each operator (I1, E1, E2, M1, M2, M3) uses this five-part template with different operator-specific instructions.
}
\label{fig:eoh_prompt_structure}
\end{figure}

\subsection{Mechanism Workflow and Iterative Search Process}
\label{app:eoh_workflow}
The EoH-style search mechanism performs program-space search through an iterative evolutionary process that combines LLM-generated program variations with fitness-based population updates.  
Algorithm~\ref{alg:eoh_br} provides the procedural view of a single best-response search, while this section summarizes the conceptual workflow.
Here, $p$, $K$, and $R$ denote user-defined hyperparameters controlling parent selection,
population size, and evolutionary depth, respectively.

At the beginning of iteration $t$ in ASRO, the search mechanism initializes a population $\mathcal{P}_0$ of candidate programs.  
If $t=0$, the population is created via prompt-based synthesis; otherwise, it is inherited from the previous ASRO iteration.  
Each program is represented by a thought--code pair and evaluated against the fixed opponent meta-strategy $\sigma_{-i}^{(t)}$.

The mechanism then runs $R$ evolutionary rounds, where $R$ is a user-defined parameter controlling the search depth.  
In each round, a subset of parents $\mathcal{P}_{\mathrm{sel}}$ is selected according to fitness and diversity.  
LLM operators (I1, E1, E2, M1, M2, M3) are instantiated as prompt templates (Figure~\ref{fig:eoh_prompt_structure}), and each operator generates new program candidates by transforming the selected parents.  
These candidates are evaluated under $\sigma_{-i}^{(t)}$, merged with the existing population, and truncated to maintain a fixed size $K$.  
After $R$ rounds, the highest-performing program in $\mathcal{P}_R$ is returned as the approximate best response.

This iterative scheme enables the search mechanism to explore both large conceptual variations (via exploration operators) and fine-grained improvements (via modification operators), while maintaining diversity across evolution.
The parameter $R$ trades optimization depth for computational cost: larger $R$ allows richer program refinement but increases total LLM calls.  
In all experiments, $R$ is kept moderate to balance performance and efficiency.

\begin{algorithm}[tb]
\caption{Program-Space Best-Response Search (LLM-Based)}
\label{alg:eoh_br}
\begin{algorithmic}[1]
\STATE \textbf{Input:} Opponent meta-strategy $\sigma_{-i}$; LLM operators $\mathcal{O}$; population size $K$; iterations $R$
\STATE \textbf{Output:} Best-performing program $p^*$

\STATE \textcolor{gray}{(0) Initialization}
\IF{first iteration}
    \STATE $\mathcal{P}^{(0)} \gets$ Prompt-based synthesis
\ELSE
    \STATE $\mathcal{P}^{(0)} \gets$ Inherit from previous population
\ENDIF

\FOR{$r = 1$ to $R$}
    \STATE \textcolor{gray}{(1) Selection}
    \STATE Select parent set $\mathcal{P}_{\mathrm{sel}} \subset \mathcal{P}^{(r-1)}$ based on fitness

    \STATE \textcolor{gray}{(2) LLM-Based Variation}
    \STATE $\mathcal{C}^{(r)} \gets \{ \textsc{LLM}(p, o) \mid p \in \mathcal{P}_{\mathrm{sel}},\, o \in \mathcal{O} \}$

    \STATE \textcolor{gray}{(3) Evaluation Against Meta-Strategy}
    \STATE Evaluate all $p \in \mathcal{C}^{(r)} \cup \mathcal{P}^{(r-1)}$ against $\sigma_{-i}$

    \STATE \textcolor{gray}{(4) Population Update}
    \STATE $\mathcal{P}^{(r)} \gets \textsc{Top-}K(\mathcal{C}^{(r)} \cup \mathcal{P}^{(r-1)})$
\ENDFOR

\STATE $p^* \gets \arg\max_{p \in \mathcal{P}^{(R)}} f(p)$
\end{algorithmic}
\end{algorithm}

\subsection{Evaluation under Mixed Strategies}
\label{app:mixed_eval}
At each ASRO iteration, candidate programs are evaluated as approximate best responses to a fixed opponent meta-strategy $\sigma_{-i}^{(t)}$.
This requires estimating expected performance under a mixed strategy rather than against a single opponent.

\textbf{Solver Evaluation.}
When evolving solver programs, the opponent strategy is the generator mixture $\sigma_g^{(t)}$.
For a solver candidate $s$, evaluation proceeds by first considering each generator $g_j$ in the current population.
A fixed set of $n_I$ instances is sampled from $g_j$, and the solver is applied to all sampled instances.
Here, $n_I$ denotes the number of problem instances sampled per generator for evaluation.
Gaps are averaged within each generator to obtain a per-generator estimate

\begin{equation*}
\widehat{\mathrm{gap}}(s, g_j)
= \frac{1}{n_I} \sum_{k=1}^{n_I} \mathrm{gap}(s, x_{j,k}),
\quad x_{j,k} \sim g_j .
\end{equation*}

where $\mathrm{gap}$ denotes the normalized reference gap defined in Eq.~\ref{eq:gap}.
The overall performance of $s$ under the mixed strategy is then computed by aggregating across generators according to $\sigma_g^{(t)}$:

\begin{equation*}
\widehat{\mathrm{gap}}(s \mid \sigma_g^{(t)})
= \sum_j \sigma_g^{(t)}(g_j)\, \widehat{\mathrm{gap}}(s, g_j).
\end{equation*}

\paragraph{Generator Evaluation.}
Generator candidates are evaluated analogously under the solver mixture $\sigma_s^{(t)}$.
For a generator $g$, all solver programs $s_\ell$ are applied to instances sampled from $g$. 
Gaps are first averaged within each solver,

\begin{equation*}
\widehat{\mathrm{gap}}(s_\ell, g)
= \frac{1}{n_I} \sum_{k=1}^{n_I} \mathrm{gap}(s_\ell, x_k),
\end{equation*}

and final performance is obtained by aggregating across solvers according to $\sigma_s^{(t)}$:

\begin{equation*}
\widehat{\mathrm{gap}}(g \mid \sigma_s^{(t)})
= \sum_{\ell} \sigma_s^{(t)}(s_\ell)\, \widehat{\mathrm{gap}}(s_\ell, g).
\end{equation*}

For generators, higher $\widehat{\mathrm{gap}}(g \mid \sigma_s^{(t)})$ indicates a stronger adversary and is therefore preferred.
In both cases, evaluation reduces to computing weighted empirical averages over solver--generator--instance interactions, where the weights are given by the opponent meta-strategy.

\section{Batch Evaluation Engine}
\label{app:batch_eval}
At each meta-game iteration of ASRO, both best-response search and payoff matrix construction require evaluating a large number of solver--generator interactions under fixed opponent mixed strategies, as defined in Section~\ref{app:mixed_eval}. These evaluations correspond to computing the mixed-strategy expectations introduced above.

Concretely, given $n_s$ solver programs, $n_g$ generator programs, and $n_I$ instances sampled per generator, a single meta-game iteration involves

\begin{equation*}
O(n_s \times n_g \times n_I)
\end{equation*}

solver executions.

To implement this evaluation workload in a unified and computationally efficient manner, we adopt a batch evaluation engine that materializes all required evaluation tasks and executes them in parallel.

\subsection{Unified Two-Stage Evaluation Pipeline}
The batch evaluation engine follows a unified two-stage pipeline consisting of task preparation and batch execution.

\textbf{Stage I: Task Preparation.}
The preparation stage constructs the complete set of evaluation tasks corresponding to solver--generator--instance triples.
First, each generator samples a fixed set of $n_I$ problem instances.
For a given generator, the same instance set is shared across all solvers to ensure consistent comparison.
If required by the gap definition, reference values (e.g., optimal solutions or bounds) are computed for all generated instances.
All solver programs are then pre-processed (e.g., compilation or validation) once prior to execution, avoiding redundant overhead during evaluation and isolating failures of individual solver programs.
Finally, evaluation tasks are constructed by enumerating all combinations.
Each task encapsulates the solver identity, generator identity, instance data, the reference value (if applicable), and the corresponding opponent meta-strategy weight.

\textbf{Stage II: Batch Task Execution.}
All prepared tasks are executed in batch.
Each task corresponds to one solver execution on a single instance and returns the observed normalized gap.
Results are grouped by solver--generator pairs and averaged across instances to produce empirical estimates of meta-game payoffs.

\subsection{Parallelization and Robustness}
\label{app:parallel}

Parallelism is applied internally within both preparation and execution stages to accelerate evaluation at scale.
Independent resource budgets are used to prevent CPU oversubscription, and all evaluation tasks are protected by timeouts.
Failures of individual solver executions are isolated and penalized without interrupting the overall evaluation process.

\section{Additional Experimental Details}
\label{app:details}
\paragraph{Hyperparameters.}
Table~\ref{tab:hyperparams} summarizes the key hyperparameters used in ASRO across domains.
The outer iteration count $T$ specifies the number of solver--generator meta-game expansions.
At each iteration, fixed-size solver and generator populations of size $K$ are maintained.
The program-space best-response oracle is implemented via an EoH-style evolutionary search, where $R_s$ and $R_g$ denote the numbers of evolutionary rounds for synthesizing solver and generator candidates, respectively.
Each candidate program is evaluated on $n_I$ instances sampled from the corresponding generator program.
To stabilize early-stage co-evolution, a fixed fraction of instances (min-ratio) is always drawn from a task-specific base generator.

In terms of LLM usage, each best-response oracle invocation synthesizes $R \times K$ candidate programs \emph{for each} EoH operator in a fixed operator set, with an additional one-time initialization at the first ASRO iteration; this procedure is applied symmetrically to both solver and generator at every iteration.

\begin{table}[t]
\centering
\small
\caption{Hyperparameter settings used in ASRO experiments across domains.}
\label{tab:hyperparams}
\begin{tabular}{lccccc}
\toprule
Task & $K$ & $R_s$ & $R_g$ & $n_I$ & min-ratio \\
\midrule
OBP  & 20 & 4 & 2 & 2 & 0.3 \\
TSP  & 10 & 2 & 2 & 3 & 0.4 \\
CVRP & 10 & 2 & 2 & 5 & 0.3 \\
\bottomrule
\end{tabular}
\end{table}

\section{Online Bin Packing (OBP)}
\label{app:OBP}
\textbf{Problem Definition.}
We consider the \emph{online bin packing problem}, where items arrive sequentially and must be assigned irrevocably to bins of fixed capacity $C$~\cite{seiden_online_2002}.
At each time step $t$, an item of size $w_t \in (0, C]$ arrives and must be placed into an existing bin with sufficient remaining capacity or into a newly opened bin.
Once placed, items cannot be moved or reassigned.
The objective is to minimize the total number of bins used.

An instance is defined by an item sequence
\[
x = (w_1, w_2, \dots, w_T),
\]
and the solver must make decisions online, without lookahead or reordering.

\subsection{Solver Program Interface.}
Each solver program defines a numerical priority function
\[
\texttt{score}(w, \mathbf{r}),
\]
which takes the current item size $w$ and a vector $\mathbf{r}$ of remaining capacities of all feasible bins, and returns a score vector aligned with $\mathbf{r}$ (one score per feasible bin).
Only the priority function varies across solvers; the decoding logic and online
constraints are fixed.

\subsection{Prompt Configuration}

We first specify task-level base prompts for online bin packing, as illustrated in Figure~\ref{fig:prompt_obp}.
These prompts define the OBP task, including the task description, the required function interfaces, and task-specific constraints for solver and generator synthesis.

During evolution, these task-level prompts are embedded into a unified prompt template (Figure~\ref{fig:eoh_prompt_structure}).
Specifically, the \emph{Task} section of the OBP prompt is placed in the \emph{Task Description} slot, defining the optimization objective and problem setting.
The function signatures specified under \emph{I/O} are enforced in the \emph{Expected Output} slot.
Detailed input/output specifications and additional constraints provided under \emph{Input/Output Info} and \emph{Other Info} are incorporated into the \emph{Note} section.
All remaining components of the prompt, including operator-specific instructions and the inclusion of parent programs, follow the standard template of the search mechanism without modification.

\begin{figure}[!t]
\centering
\caption{Base prompt examples for solver and generator synthesis in online bin packing (OBP).}
\label{fig:prompt_obp}
\begin{tcolorbox}[
  enhanced,
  colback=promptbg,
  colframe=promptframe,
  boxrule=0.6pt,
  arc=10pt,
  fonttitle=\bfseries\color{prompttitle},
  width=\linewidth,
  sidebyside,
  sidebyside gap=12pt,
  equal height group=obpPrompts,
  righthand width=0.4485\linewidth
]

\small
\RaggedRight

\textbf{Prompt for Generator}

\vspace{0.4em}
\textbf{Task.}\\
Write a novel Python function to generate sequences of items for the online bin packing problem.
Goal: Create instances that are as difficult as possible for heuristic solvers to pack efficiently.
The difficulty must arise from the item arrival order and the strategic relationship between sequential items.

\vspace{0.55em}
\textbf{I/O.}\\
Implement \texttt{generate\_instances(seeds, capacity, num\_items)} and return \texttt{instances}.

\vspace{0.55em}
\textbf{Input/Output Info.}\\
\texttt{seeds}: iterable of integers, one per instance.\\
\texttt{capacity}: bin capacity (int $>1$).\\
\texttt{num\_items}: number of items in each instance (int $>0$).\\
\texttt{instances}: Python list with the same length as seeds; each element is a NumPy integer array named \texttt{items} of shape $(num\_items,) $with values strictly in $[1, capacity-1]$.

\vspace{0.55em}
\textbf{Other Info.}\\
Avoid simple random noise; instead, design structured sequences that force poor decision-making in real-time assignments.
Avoid trivial or degenerate streams that make the objective uninformative.
Output ONLY the function code.

\tcblower
\small
\RaggedRight

\textbf{Prompt for Solver}

\vspace{0.4em}
\textbf{Task.}\\
Write a novel scoring Python function for the online bin packing problem using a greedy strategy.
Goal: Minimize the total number of bins by making each assignment facilitate future packing.
Strategy Hint: Consider not just the tightest fit, but the potential usability of the remaining space. A good score should balance between filling a bin and avoiding leaving behind tiny, unusable gaps that might reject future items. 

\vspace{0.55em}
\textbf{I/O.}\\
Implement \texttt{score(item, bins)} and return \texttt{scores}.

\vspace{0.55em}
\textbf{Input/Output Info.}\\
\texttt{item}: current item size.\\
\texttt{bins}: NumPy array of remaining capacities for \emph{feasible bins only} (each entry satisfies $bins[i] >= item$).\\
\texttt{scores}: NumPy array with the same shape as \texttt{bins}. Higher is better.\\

\vspace{0.55em}
\textbf{Other Info.}\\
Be scale-invariant (use ratios such as $(bins-item)/max(bins)$ or statistics of \texttt{bins} instead of hard-coded thresholds).
Output ONLY the function code.

\end{tcolorbox}
\end{figure}

\subsection{Evolution Outcomes}
\begin{figure}[!t]
    \centering
    \includegraphics[width=1\linewidth]{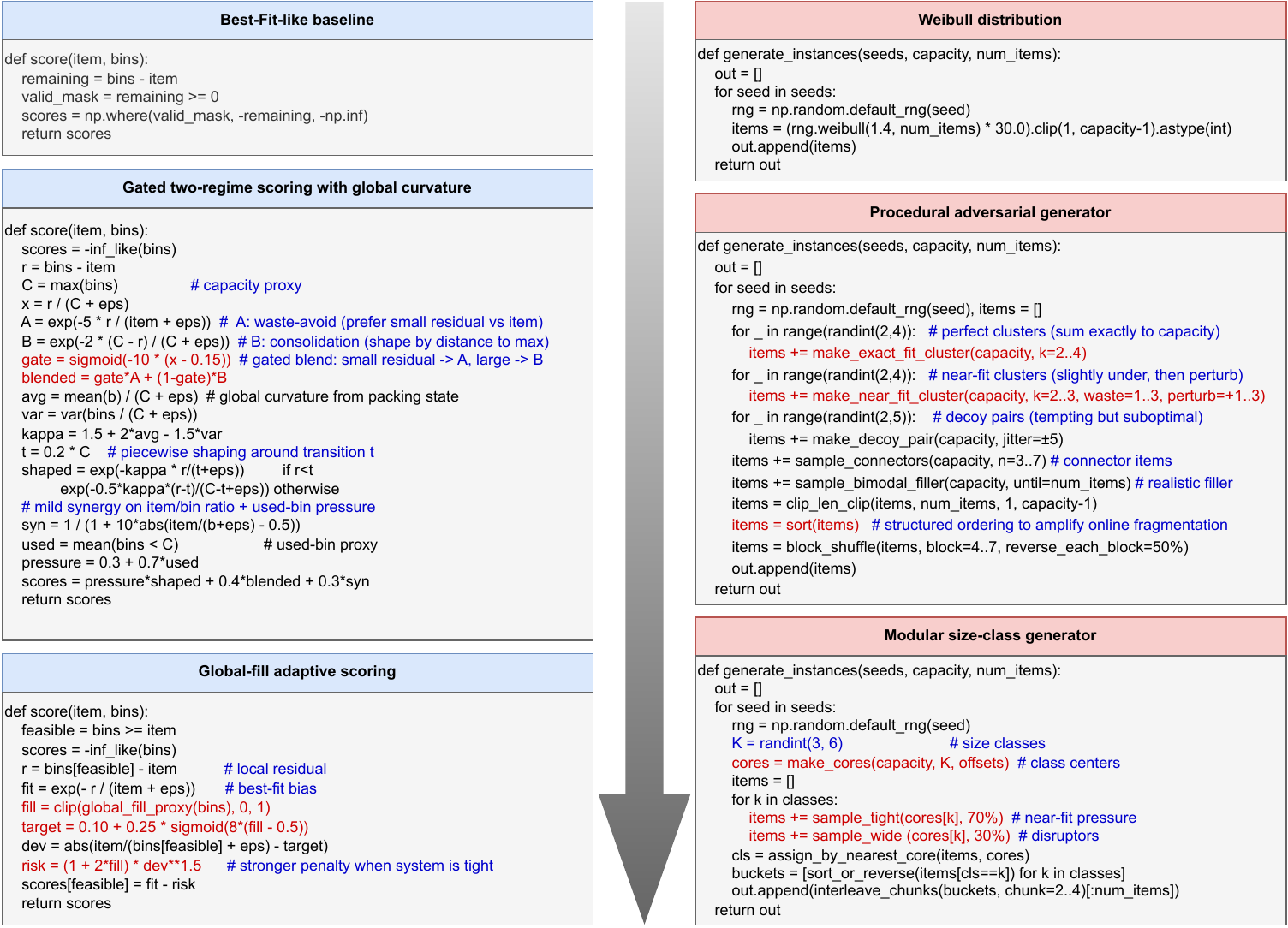}
    \caption{\textbf{Representative solver and generator programs discovered by ASRO on online bin packing.}
    The figure shows representative solver scoring rules (left) and instance generators (right) selected from the final solver and generator pools.
    Code is schematized to emphasize algorithmic structure rather than implementation details.}
    \label{fig:obp_programs}
\end{figure}

Figure~\ref{fig:obp_programs} provides a qualitative snapshot of the solver--generator strategy space shaped by co-evolution in ASRO.
The figure visualizes representative executable programs selected from the final solver and generator pools, highlighting characteristic structural patterns induced by sustained solver--generator interaction.
On the solver side, co-evolution with adversarial instance generators gives rise to scoring rules that extend beyond purely local heuristics by incorporating global packing statistics and adaptive penalty mechanisms.
On the generator side, pressure from increasingly capable solvers leads to structured instance constructions that induce systematic pairing conflicts and amplify online misallocation.
Together, these examples illustrate how solver and generator strategies mutually adapt under ASRO, resulting in a diverse and interdependent strategy space.

\subsection{Datasets and Arrival Orders.}
We evaluate OBP solvers on a collection of widely used public benchmarks as well as synthetic distributions.
Specifically, we use the Falkenauer \texttt{T} and \texttt{U} benchmark families~\cite{falkenauer_hybrid_1996}, the \texttt{Hard28} benchmark~\cite{delorme_bpplib_2018}. We additionally evaluate on synthetic instances generated from Weibull distributions, as commonly used in online bin packing studies~\cite{castineiras_weibullbased_2012}.
These datasets cover both classical test cases and structurally challenging scenarios commonly studied in the online bin packing literature.

Since online bin packing performance is sensitive to item arrival order~\cite{albers_best_2020}, we evaluate all methods under multiple arrival models.
In particular, we consider random order and size-based orders (ascending).
Arrival orders are fixed independently of the evaluated algorithms and applied consistently across all methods to ensure fair comparison.

\subsection{Baselines.}
We compare ASRO-generated solvers against classical online bin packing heuristics and the EoH baseline.
The classical baselines include \emph{First Fit}~\cite{Ullman1971performance, dosa_first_2013} and \emph{Best Fit}~\cite{johnson_worstcase_1974, albers_best_2020}, which represent standard greedy strategies widely used in practice.
In addition, we include the solver evolved by \emph{Evolution of Heuristics (EoH)}~\cite{liu_evolution_2024}, which follows a static, solver-centric LLM-based heuristic discovery pipeline without adversarial instance generation.
All baselines operate under the same online constraints and decoding procedure as ASRO solvers, without lookahead, reordering, or post-processing.

\section{Traveling Salesman Problem (TSP)}
\label{app:TSP}
\subsection{Problem Definition.}
We consider the Euclidean Traveling Salesman Problem (TSP) with $N$ cities.
Given city coordinates $\{c_i\}_{i=1}^N \subset \mathbb{R}^2$, the objective is to find a Hamiltonian tour of minimum total length.
Equivalently, for a permutation $\pi$ of $\{1,\dots,N\}$, the tour cost is
\[
V(\pi) \;=\; \sum_{i=1}^{N} d\!\left(c_{\pi_i}, c_{\pi_{i+1}}\right),
\quad \text{with } \pi_{N+1}=\pi_1,
\]
where $d(\cdot,\cdot)$ is the Euclidean distance.
The solver aims to minimize $V(\pi)$.

\subsection{Solver Procedure and Program Interface (TSP)}
\paragraph{Solver Procedure (Guided Local Search).}
For the Traveling Salesman Problem (TSP), we adopt Guided Local Search (GLS)~\cite{voudouris_guided_1999a, alsheddy_guided_2018,arnold_knowledgeguided_2019} as the fixed solver procedure.
GLS is a local-search-based algorithm that iteratively improves a candidate tour using a predefined set of neighborhood operators, specifically 2-opt and relocate.
At any point in the search, candidate local moves are evaluated under an
\emph{augmented edge-distance matrix} $D = (d_{ij})$,
which is initialized from the Euclidean distances and subsequently modified
by solver-defined update rules to incorporate adaptive edge penalties.

Starting from an initial tour, GLS repeatedly applies local search operators under $D$ until a locally optimal tour $\pi^{\mathrm{loc}}$ is reached, i.e., no improving local move exists under $D$.
Based on $\pi^{\mathrm{loc}}$, GLS updates an edge-usage matrix $U = (u_{ij})$, 
where $u_{ij}$ records how frequently edge $(i,j)$ has been selected in locally optimal tours during the search history.
The search then continues by updating the augmented edge-distance matrix
via the solver-defined update rule and resuming local search under the updated distances.

\paragraph{Solver Program Interface.}
Within this fixed GLS procedure, solver programs specify an \emph{edge-distance update rule} that  modifies the augmented edge-distance matrix.
Given the current augmented edge-distance matrix $D$, the current locally optimal tour $\pi^{\mathrm{loc}}$, and the edge-usage matrix $U$, the solver program outputs an updated augmented edge-distance matrix:
\[
\tilde{D} = \texttt{update\_edge\_distance}(D, \pi^{\mathrm{loc}}, U).
\]

The updated matrix $\tilde{D}$ is then used by GLS to evaluate subsequent local moves.
All other components of the solver procedure, including the local search operators and the GLS control logic, are fixed across methods; only the distance-update rule varies across solver programs.

\subsection{Prompt Configuration}
We specify task-level base prompts for the traveling salesman problem, as illustrated in Figure~\ref{fig:prompt_tsp}.
These prompts define the TSP task, including the optimization objective, required function interfaces, and task-specific constraints for solver and generator synthesis.
Prompt construction for TSP follows the same procedure as in online bin packing.
Specifically, the TSP base prompts are embedded into the unified prompt template structure shown in Figure~\ref{fig:eoh_prompt_structure}, with task-specific content populated accordingly and all remaining components following the standard template of the search mechanism.

\begin{figure}[t]
\centering
\caption{Base prompt examples for solver and generator synthesis in Traveling Salesman Problem (TSP).}
\label{fig:prompt_tsp}
\begin{tcolorbox}[
  enhanced,
  colback=promptbg,
  colframe=promptframe,
  boxrule=0.6pt,
  arc=10pt,
  fonttitle=\bfseries\color{prompttitle},
  width=\linewidth,
  sidebyside,
  sidebyside gap=12pt,
  equal height group=tspPrompts,
  righthand width=0.52\linewidth,
  fontupper=\small,      
  fontlower=\small      
]

\RaggedRight
\textbf{Prompt for Generator}

\vspace{0.4em}
\textbf{Task.}\\
Write a novel Python function to generate Euclidean Traveling Salesman Problem instances.
The goal is to create instances that are as difficult as possible for heuristic solvers,
measured by the resulting tour length.
The difficulty should arise from the geometric structure of city coordinates and their spatial arrangement,
rather than from simple random noise.
Design instances that induce misleading local structures, long-range dependencies,
or deceptive spatial patterns that make greedy or local-search-based solvers struggle.

\vspace{0.55em}
\textbf{I/O.}\\
Implement \texttt{generate\_instances(seeds, n\_cities)} and return \texttt{instances}.

\vspace{0.55em}
\textbf{Input/Output Info.}\par
\texttt{seeds}: iterable of integers, one per instance. \\
\texttt{n\_cities}: number of cities in each TSP instance. \\
\texttt{instances}: Python list with the same length as \texttt{seeds};
each element is a NumPy array of shape $(n\_cities, 2)$,
representing city coordinates in \texttt{$[0,1]^2$}.\par

\vspace{0.55em}
\textbf{Other Info.}\\

Output ONLY the function code.

\tcblower
\RaggedRight
\textbf{Prompt for Solver}

\vspace{0.4em}
\textbf{Task.}\\
Write a GLS-style edge-distance update rule as a Python function that helps the solver minimize total tour length.
The update rule should modify edge penalties based on the structure of a locally optimized tour, encouraging exploration of alternative edges and discouraging repeatedly used suboptimal edges.
The goal is to improve solution quality across the sampled instance distribution, not just on a single instance.

\vspace{0.55em}
\textbf{I/O.}\\
Implement \texttt{update\_edge\_distance(edge\_distance, local\_opt\_tour, edge\_n\_used)}
and return \texttt{updated\_edge\_distance}.

\vspace{0.55em}
\textbf{Input/Output Info.}\\
\texttt{edge\_distance}: NumPy array of shape $(N, N)$, symmetric.\\
\texttt{local\_opt\_tour}: list or 1D NumPy array of city indices representing a tour.\\
\texttt{edge\_n\_used}: NumPy array of shape $(N, N)$, symmetric, storing edge usage counts.\\
\texttt{updated\_edge\_distance}: NumPy array of shape $(N, N)$, symmetric.

\vspace{0.55em}
\textbf{Other Info.}\\
Do NOT modify input arrays in-place; operate on copies and return updated copies.
Output ONLY the function code.
\end{tcolorbox}
\end{figure}

\subsection{Evolution Outcomes}
\begin{figure}[!t]
    \centering
    \includegraphics[width=0.9\linewidth]{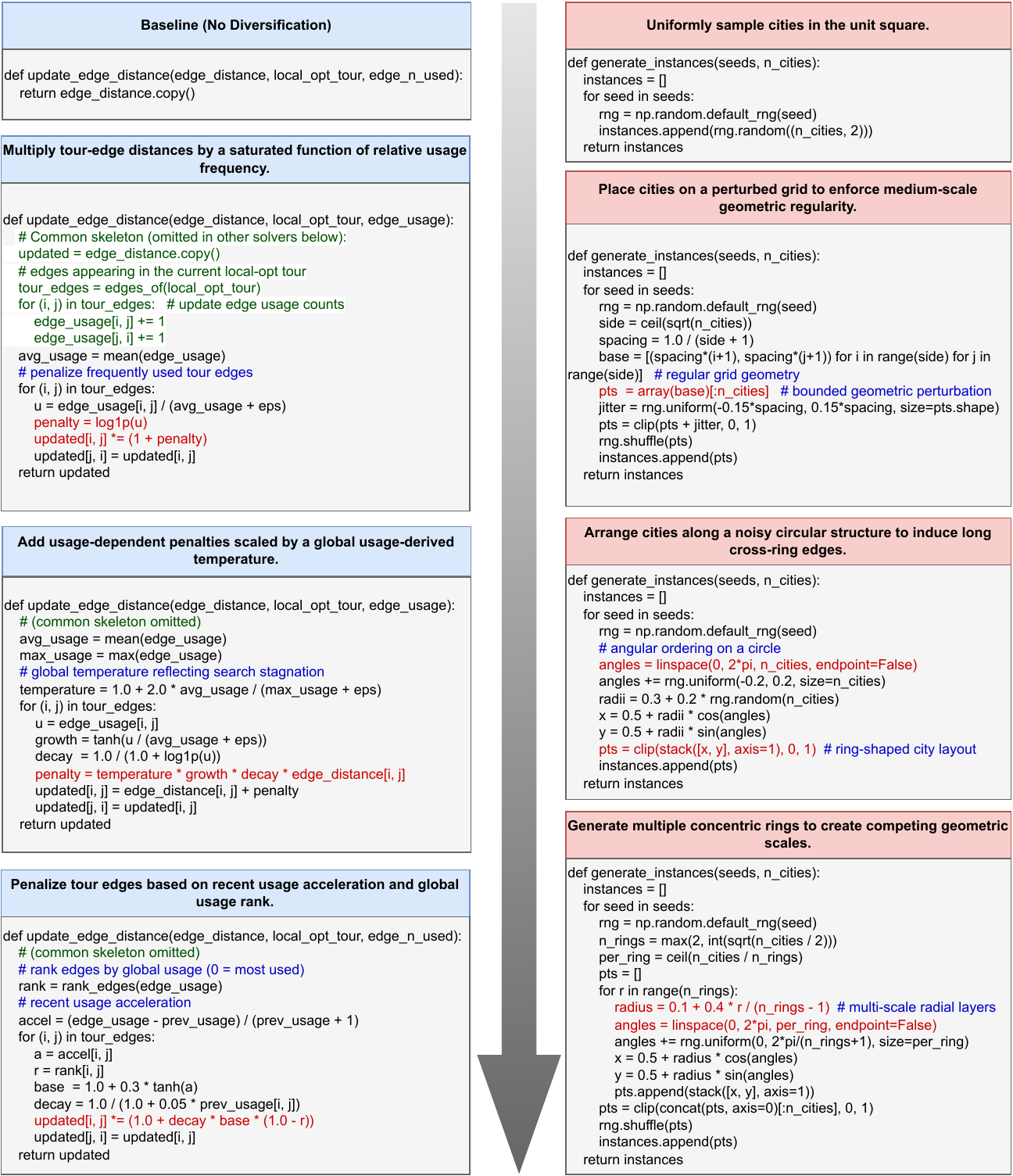}
    \caption{\textbf{Representative solver and generator programs discovered by ASRO on the Traveling Salesman Problem.}
    We show distilled solver distance-update rules (left) and instance generators (right) drawn from the final solver and generator pools.
    Code is schematized for clarity to emphasize algorithmic structure; auxiliary implementation details are omitted.}
    \label{fig:tsp_programs}
\end{figure}

Figure~\ref{fig:tsp_programs} shows representative solver and generator programs learned by ASRO on the Traveling Salesman Problem. Solver programs evolve toward more structured distance-update mechanisms, while instance generators progressively induce structured geometries that interact with these updates.

\subsection{Benchmark Datasets}
We evaluate the learned TSP solvers on benchmark instances from \emph{TSPLIB}~\cite{reinelt_tsplib_1991}, a standard library of Traveling Salesman Problem instances widely used in the combinatorial optimization literature.
TSPLIB instances exhibit substantial diversity in both scale and geometric structure, including uniform and clustered point sets, grid-like layouts, and instances with irregular spatial patterns.

The benchmark covers a broad range of problem sizes, from small instances with fewer than 200 cities to large-scale instances exceeding 1{,}000 cities.
All TSPLIB instances considered are evaluated against reference tour lengths (e.g., best-known solutions), enabling standardized reporting via normalized optimality gaps.

Importantly, TSPLIB instances are not drawn from the distributions used during training.
Evaluating on TSPLIB therefore provides a stringent test of generalization under distributional and structural shift.

\subsection{More Results: Per-instance TSPLIB Performance}
Table~\ref{tab:tsplib_per_instance} reports per-instance results on TSPLIB benchmarks, comparing the final ASRO solver against the EoH baseline.
Rather than aggregating by instance size or family, this table exposes instance-level behavior and highlights where improvements are consistently realized.

\begin{table}[t]
\centering
\caption{Per-instance TSPLIB results. Lower is better.}
\label{tab:tsplib_12cols}
\setlength{\tabcolsep}{3.8pt}
\renewcommand{\arraystretch}{1.0}
\scriptsize

\begin{tabular}{lrr lrr lrr lrr}
\toprule
Instance & ASRO-EoH & EoH &
Instance & ASRO-EoH & EoH &
Instance & ASRO-EoH & EoH &
Instance & ASRO-EoH & EoH \\
\cmidrule(lr){1-3}\cmidrule(lr){4-6}\cmidrule(lr){7-9}\cmidrule(lr){10-12}

A280     & 0.301 & 2.195 &
FNL4461  & 4.635 & 4.635 &
PCB442   & 0.462 & 0.990 &
RD400    & 0.163 & 2.717 \\

BERLIN52 & 0.031 & 2.279 &
GIL262   & 0.414 & 0.955 &
PR1002  & 2.245 & 3.051 &
RL1304  & 3.780 & 6.771 \\

BIER127  & 0.318 & 0.224 &
KROA100  & 0.016 & 0.119 &
PR107   & -0.003 & 0.614 &
RL1323  & 2.582 & 4.764 \\

CH130    & 0.012 & 0.381 &
KROA150  & 0.003 & 0.639 &
PR124   & 0.001 & 0.096 &
RL1889  & 2.784 & 3.879 \\

CH150    & 0.044 & 1.282 &
KROA200  & 0.005 & 0.907 &
PR136   & -0.001 & 0.459 &
ST70    & 0.313 & 0.313 \\

D1291    & 3.649 & 5.484 &
KROB100  & -0.009 & -0.009 &
PR144   & -0.003 & 0.353 &
TS225   & 2.582 & 4.764 \\

D1655    & 4.006 & 4.144 &
KROB150  & -0.009 & 1.367 &
PR152   & 0.002 & 0.907 &
TSP225  & -0.901 & 0.561 \\

D198     & 0.301 & 1.056 &
KROB200  & 0.015 & 0.819 &
PR226   & 0.002 & 0.877 &
U1060   & 1.850 & 4.132 \\

D2103    & 1.068 & 1.839 &
KROC100  & 0.008 & 0.008 &
PR2392  & 4.186 & 4.186 &
U1432   & 2.645 & 4.069 \\

D493     & 1.249 & 2.998 &
KROD100  & 0.001 & 0.317 &
PR264   & 0.000 & 0.231 &
U159    & -0.010 & -0.010 \\

D657     & 1.012 & 3.023 &
KROE100  & 0.024 & 0.438 &
PR299   & 0.039 & 0.791 &
U1817   & 4.072 & 4.168 \\

EIL101   & 1.782 & 4.144 &
LIN105   & 0.028 & 0.028 &
PR439   & 2.255 & 5.300 &
U2152   & 4.025 & 4.491 \\

EIL51    & 0.674 & 0.700 &
LIN318   & 0.342 & 3.738 &
PR76    & 0.000 & 1.467 &
U2319   & 2.366 & 2.366 \\

EIL76    & 1.184 & 1.732 &
LINHP318 & 2.002 & 5.454 &
RAT195  & 0.582 & 0.982 &
U574    & 1.174 & 5.071 \\

FL1400   & 1.824 & 5.096 &
NRW1379  & 2.829 & 3.961 &
RAT575  & 2.183 & 3.152 &
U724    & 1.411 & 3.161 \\

FL1577   & 2.409 & 5.031 &
P654     & 0.404 & 2.080 &
RAT783  & 2.363 & 3.907 &
VM1084  & 1.158 & 3.552 \\

FL3795   & 4.382 & 4.382 &
PCB1173  & 3.246 & 5.004 &
RAT99   & 0.681 & 0.681 &
VM1748  & 2.133 & 3.280 \\

FL417    & 0.509 & 0.984 &
PCB3038  & 4.132 & 4.132 &
RD100   & 0.005 & 0.016 &
         &       &       \\

\bottomrule
\end{tabular}
\label{tab:tsplib_per_instance}
\end{table}

\subsection{Baselines}
\emph{Classical constructive heuristics} include Nearest Insertion (NI) and Farthest Insertion (FI)~\cite{applegate_traveling_2006, rosenkrantz_analysis_2009}, which build tours incrementally according to distance-based criteria.
These methods are simple, deterministic, and widely used as standard non-learning baselines.

\section{Capacitated Vehicle Routing Problem (CVRP)}
\label{app:CVRP}
\subsection{Problem Definition.}
We consider the Capacitated Vehicle Routing Problem (CVRP)~\cite{toth_vehicle_2014}.
A fleet of identical vehicles departs from and returns to a single depot to serve a set of customers with known demands.
Each customer must be visited exactly once, and the total demand served by any vehicle cannot exceed the vehicle capacity $Q$.
The objective is to minimize the total travel cost over all routes.
An instance is defined by a depot location, customer locations $\{c_i\}_{i=1}^N$, customer demands $\{d_i\}_{i=1}^N$, and a vehicle capacity $Q$.
Feasibility with respect to capacity constraints must be satisfied at all times.

\subsection{Solver Program Interface.}
Solver programs for CVRP are implemented as step-wise selection functions
invoked within a fixed greedy route-construction procedure.
At each decision step, given the current node $v \in \{0,1,\dots,N\}$
(with $v=0$ denoting the depot),
the depot index $v_0 = 0$,
the feasible unvisited customer index collection $U$ (a subset of $\{1,\dots,N\}$),
the remaining vehicle capacity $q$,
the customer demand vector $\mathbf{d} \in \mathbb{R}^{N+1}$ (with $d_0 = 0$),
and the distance matrix $D \in \mathbb{R}^{(N+1)\times(N+1)}$,
the solver program selects the next node to visit:
\[
i^\star = \texttt{select}(v, v_0, U, q, \mathbf{d}, D).
\]

The decoding procedure enforces feasibility by restricting $U$
to customers whose demand does not exceed the remaining capacity $q$.
If no feasible customer exists, the current route is terminated and
the vehicle returns to the depot.
Only the selection function varies across solver programs;
all other components of the route-construction procedure are fixed.

\subsection{Prompt Configuration}
We specify task-level base prompts for the capacitated vehicle routing problem, as illustrated in Figure~\ref{fig:prompt_cvrp}.
These prompts define the CVRP task, including the optimization objective, required function interfaces, and task-specific constraints for solver and generator synthesis.
Prompt construction for CVRP follows the same procedure as in online bin packing.
Specifically, the CVRP base prompts are embedded into the unified prompt template structure shown in Figure~\ref{fig:eoh_prompt_structure}, with task-specific content populated accordingly and all remaining components following the standard template of the search mechanism.

\begin{figure}[t]
\centering
\caption{Base prompt examples for solver and generator synthesis in the Capacitated Vehicle Routing Problem (CVRP).}
\label{fig:prompt_cvrp}
\begin{tcolorbox}[
  enhanced,
  colback=promptbg,
  colframe=promptframe,
  boxrule=0.6pt,
  arc=10pt,
  fonttitle=\bfseries\color{prompttitle},
  width=\linewidth,
  sidebyside,
  sidebyside gap=12pt,
  equal height group=cvrpPrompts,
  righthand width=0.36\linewidth
]

\small
\RaggedRight
\textbf{Prompt for Generator}

\vspace{0.4em}
\textbf{Task.}\\
Write a novel Python function that produces realistic, well-structured CVRP instances designed to reduce the expected solution quality achieved by a mixed pool of heuristic solvers.
The difficulty should arise from general structural properties of customer distributions, demand patterns, and capacity constraints, rather than solver-specific tricks.

The generation process may follow a structured protocol:
(A) choose a sampling plan that induces hard solution landscapes (e.g., clustered customers,
uneven demands, tight capacity);
(B) sample realistic customer locations, avoiding extreme clustering or purely uniform layouts;
(C) assign demands that create non-trivial capacity challenges via a mix of small and large demands;
(D) perform lightweight self-checks (e.g., demand sum, capacity feasibility, spatial spread),
and resample with bounded attempts if degenerate cases arise.

\vspace{0.55em}
\textbf{I/O.}\\
Implement \texttt{generate\_instances(seeds, num\_customers, vehicle\_capacity)}
and return \texttt{instances}.

\vspace{0.55em}
\textbf{Input/Output Info.}\\
\texttt{seeds}: iterable of integers, one per instance.\\
\texttt{num\_customers}: number of customers per instance (int).\\
\texttt{vehicle\_capacity}: vehicle capacity (int).\\
\texttt{instances}: Python list with the same length as \texttt{seeds}; each element is a dict with
exactly the following keys: \texttt{depot}: list \texttt{[x, y]} representing depot coordinates; \texttt{customers}: list of dicts, each with \texttt{'coords'} \texttt{[x, y]} in $[0,100] \times [0,100]$ and \texttt{'demand'} (positive integer); \texttt{vehicle\_capacity}: integer.

\vspace{0.55em}
\textbf{Other Info.}\\
Avoid pathological or degenerate instances; prefer bounded resampling with quick checks over heavy computation.
Output ONLY the Python function code.

\tcblower

\small
\RaggedRight
\textbf{Prompt for Solver}

\vspace{0.4em}
\textbf{Task.}\\
Write a step-by-step Python construction heuristic for the CVRP that selects the next node at each step to minimize total travel distance while respecting vehicle capacity constraints.
The heuristic should aim to perform well across instances drawn from a mixture of generators,
rather than overfitting to a single instance distribution.

\vspace{0.55em}
\textbf{I/O.}\\
Implement \texttt{select(current\_node, depot, unvisited\_nodes, rest\_capacity, demands, distance\_matrix)}
and return \texttt{next\_node}.

\vspace{0.55em}
\textbf{Input/Output Info.}\\
\texttt{current\_node}: int, current location ($0$ denotes the depot; $1..n$ are customers).\\
\texttt{depot}: int, always $0$.\\
\texttt{unvisited\_nodes}: NumPy array of feasible customer IDs (subset of $\{1,\dots,n\}$).\\
\texttt{rest\_capacity}: remaining vehicle capacity (float).\\
\texttt{demands}: NumPy array of length $(n+1)$ with \texttt{demands[0]=0}.\\
\texttt{distance\_matrix}: NumPy array of shape $(n+1, n+1)$ with symmetric distances.\\
\texttt{next\_node}: int, either $0$ (return to depot) or an element of \texttt{unvisited\_nodes}.

\vspace{0.55em}
\textbf{Other Info.}\\
Output ONLY the Python function code.

\end{tcolorbox}
\end{figure}

\subsection{Evolution Outcomes}
\begin{figure}[!t]
    \centering
    \includegraphics[width=1\linewidth]{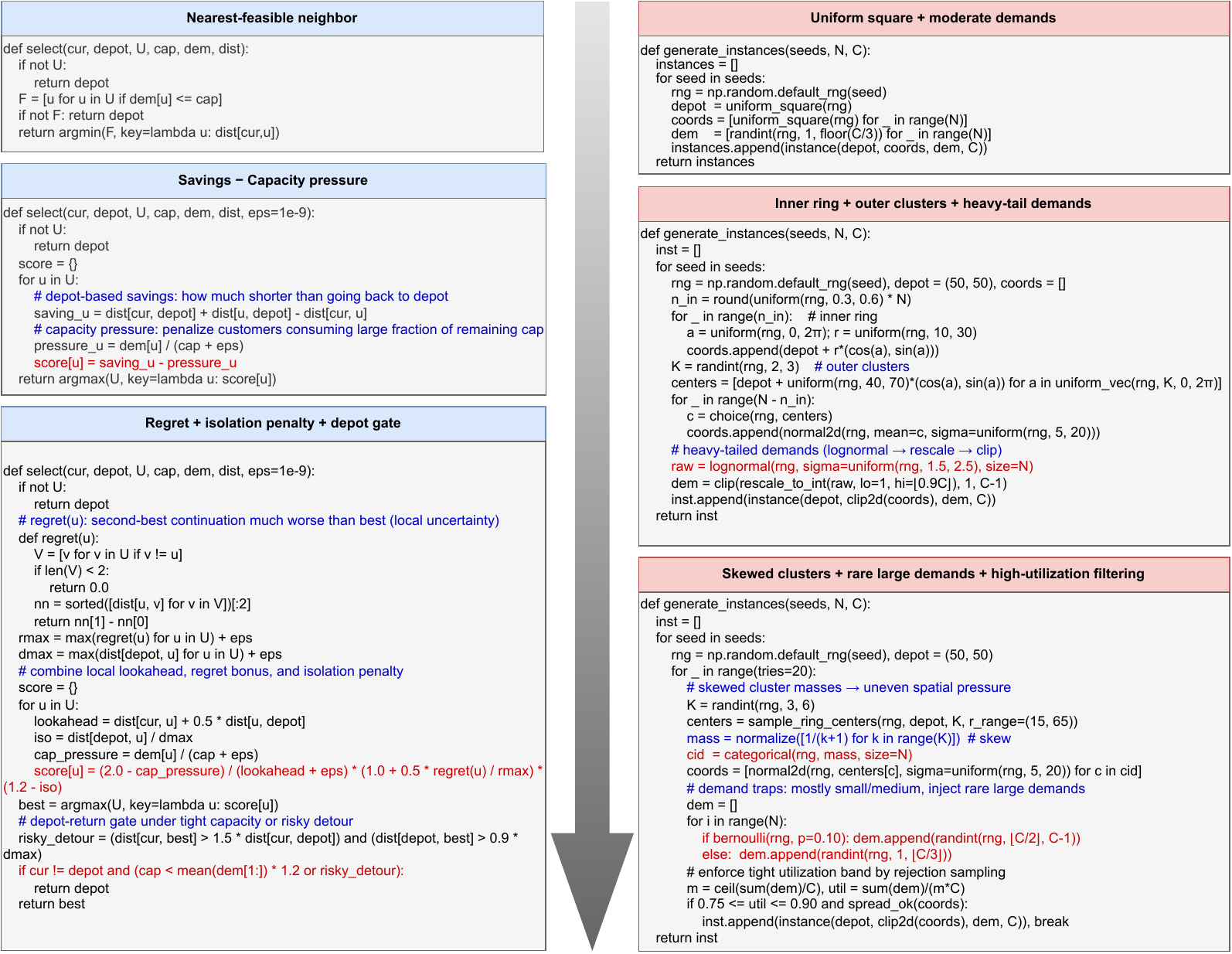}
    \caption{\textbf{Representative solver and generator programs discovered by ASRO on the capacitated vehicle routing problem (CVRP).}
   We show distilled solver customer-selection rules (left) and instance generators (right) drawn from the final solver and generator pools.
    Code is schematized to highlight the core algorithmic logic induced by adversarial pressure; auxiliary implementation details (e.g., batching, feasibility filtering, and acceptance criteria) are omitted for clarity.}
    \label{fig:cvrp_programs}
\end{figure}

Figure~\ref{fig:cvrp_programs} visualizes representative solver and generator programs discovered by ASRO on CVRP.
On the solver side, the learned rules incorporate savings- and regret-shaped customer selection under explicit capacity pressure.
On the generator side, instance constructions evolve from simple sampling to spatially clustered geometries with heterogeneous demand profiles that stress routing decisions.

\subsection{Benchmarks (CVRPLIB).}
We evaluate CVRP solvers on benchmark instances from \emph{CVRPLIB}~\cite{uchoa_new_2017}, a widely used public repository of capacitated vehicle routing problems.
CVRPLIB contains instances with diverse spatial distributions, demand patterns, and capacity constraints, and is commonly adopted for evaluating both heuristic and learning-based routing algorithms.

We consider instance families \texttt{A}, \texttt{B}, \texttt{E}, \texttt{F}, \texttt{M}, \texttt{P}, and \texttt{X}.
These families span a broad range of problem sizes and structural characteristics~\cite{uchoa_new_2017}.
In particular, \texttt{A} and \texttt{B} instances are relatively regular and homogeneous, while \texttt{E}, \texttt{F}, \texttt{M}, \texttt{P}, and \texttt{X} exhibit increasing heterogeneity in customer distributions, route lengths, and demand structures.

All CVRPLIB instances used for evaluation are distinct from the synthetic distributions employed during training.
Performance is measured using normalized optimality gaps with respect to reference solutions, providing a standardized and widely accepted evaluation protocol.

\section{Additional Ablation Results}
\label{app:ablation}
This appendix reports detailed ablation results that support the analysis in Section~\ref{sec:ablation}.
Due to space constraints, these results are omitted from the main text.

\subsection{Data Augmentation and Self-play}
\paragraph{TSP Data augmentation distribution.}
\label{app:tsp_da}
The data augmentation (DA) training distribution is constructed as a fixed mixture of five geometric families with weights: uniform square (30\%), clustered Gaussian (25\%), grid-jitter (15\%), annulus/ring (15\%), and two-scale mixture (15\%).
For each instance, a family is sampled according to these weights, followed by family-specific coordinate generation.
Full specifications are given below.

\textbf{(A) Uniform square (30\%).}
Each city is sampled independently as $(x,y) \sim \mathrm{Unif}([0,1]^2)$.

\textbf{(B) Clustered Gaussian (25\%).}
The number of clusters is sampled as $k \sim \mathrm{Unif}\{3,\ldots,8\}$, with cluster centers drawn uniformly in $[0,1]^2$. Each city is assigned to a random cluster and sampled as $(x,y) = c_j + \epsilon$, where $\epsilon \sim \mathcal{N}(0,\sigma^2 I)$ and
$\sigma \sim \mathrm{Unif}[0.02,0.08]$. Coordinates are clipped to $[0,1]^2$.

\textbf{(C) Grid-jitter (15\%).}
An approximately $\lceil \sqrt{n} \rceil \times \lceil \sqrt{n} \rceil$ regular grid is constructed over $[0,1]^2$, where $n$ denotes the number of cities. Grid points are perturbed by i.i.d.\ noise $\delta_x,\delta_y \sim \mathrm{Unif}[-0.05,0.05]$, and $n$ points are selected; if necessary, remaining points are sampled uniformly.

\textbf{(D) Annulus / ring (15\%).}
Angles are sampled uniformly in $[0,2\pi]$ and radii are sampled uniformly from $[r_0,r_1]$, where $r_0 \sim \mathrm{Unif}[0.2,0.4]$ and $r_1 \sim \mathrm{Unif}[0.5,0.7]$.
Polar coordinates are converted to Cartesian coordinates, translated to the unit-square center, and clipped to $[0,1]^2$.

\textbf{(E) Two-scale mixture (15\%).}
Seventy percent of cities are sampled uniformly over $[0,1]^2$, while the remaining 30\% are sampled within a small square of side length sampled from $[0.05,0.15]$ at a random corner. City order is randomly permuted after generation.

\paragraph{Self-play baseline.}
The self-play variant removes the meta-game structure of ASRO while retaining alternating solver--generator optimization.
At each iteration, the solver and generator are updated only against their most recent counterparts, without maintaining a strategy pool or computing mixed meta-strategies.
Concretely, self-play is implemented by restricting both the solver and generator strategy pools to size one, so that only the most recent program is retained on each side.
This results in a greedy, memoryless adversarial loop rather than equilibrium-based co-evolution.

Table~\ref{tab:tsp_ablation_da} reports the corresponding ablation results.
\begin{table}[t]
\centering
\caption{\textbf{Ablation on training strategies for TSP.}
Comparison between EoH with data augmentation (DA) or self-play (SP) and ASRO-EoH.
Results are reported as average optimality gaps (\%$\downarrow$).}
\renewcommand{\arraystretch}{1}
\resizebox{.48\textwidth}{!}{%
\begin{tabular}{@{\hspace{1mm}}lccccc@{\hspace{1mm}}}
\toprule
Method & LIB-S & LIB-M & LIB-L & LIB-XL & Uniform \\
\midrule
EoH
& 0.77 & 1.99 & 3.40 & 4.20 & 0.27 \\
EoH + SP
& 0.30 & 1.06 & 2.22 & 3.50 & 0.22 \\
EoH + DA
& 0.44 & 1.17 & 2.03 & 3.66 & 0.28 \\
\midrule
\rowcolor{lightgray!40}
ASRO-EoH
& \textbf{0.23} & \textbf{0.67} & \textbf{1.73} & \textbf{2.82} & \textbf{0.06} \\
\bottomrule
\end{tabular}%
}
\label{tab:tsp_ablation_da}
\end{table}

\subsection{Effect of LLM Backbones}
\paragraph{Backbone specification.}
We evaluate ASRO with multiple LLM backbones for program-space search: DeepSeek-V3.2~\cite{deepseek-ai_deepseekv32_2025}, Gemini-3-Flash-Preview, GPT-5-Nano, GPT-5-Mini, and LLaMA-3-8B-Instruct~\cite{grattafiori_llama_2024}.
All models are used with identical prompts and decoding configurations.
Table~\ref{tab:tsp_ablation_llm} reports the corresponding ablation results.
\begin{table}[t]
\centering
\caption{\textbf{Effect of different LLM backbones in ASRO (TSP).}
All variants use the same ASRO-EoH framework and differ only in the LLM used for program-space search; the \textbf{EoH} baseline (DeepSeek) is included for reference.
Results are reported as average optimality gaps (\%$\downarrow$).}
\renewcommand{\arraystretch}{1}
\resizebox{.48\textwidth}{!}{%
\begin{tabular}{@{\hspace{1mm}}lccccc@{\hspace{1mm}}}
\toprule
LLM & LIB-S & LIB-M & LIB-L & LIB-XL & Uniform \\
\midrule
DeepSeek
& 0.23 & 0.67 & 1.73 & 2.82 & 0.06 \\
Gemini
& 0.23 & \textbf{0.45} & \textbf{0.96} & \textbf{2.45} & \textbf{0.06} \\
GPT-5 nano
& 0.26 & 0.69 & 2.38 & 3.68 & 0.10 \\
GPT-5 mini
& \textbf{0.21} & 0.47 & 1.39 & 2.53 & 0.10 \\
LLaMA-3-8B
& 0.31 & 1.01 & 3.00 & 4.33 & 0.20 \\
\midrule
EoH (DeepSeek)
& 0.77 & 1.99 & 3.40 & 4.20 & 0.27 \\

\bottomrule
\end{tabular}%
}

\label{tab:tsp_ablation_llm}
\end{table}

\subsection{Oracle-agnosticity}

ReEvo~\cite{ye_reevo_2024} is an evolutionary program-synthesis framework that integrates LLMs into evolutionary search by using LLMs both to generate new heuristic candidates and to produce reflective feedback that guides evolution.
Compared to the EoH-style program search mechanism, which relies on implicit feedback from heuristic evaluations, ReEvo explicitly incorporates reflective LLM feedback into the evolutionary loop via short- and long-term reflections that guide crossover and mutation.
In this ablation, the EoH-style program search mechanism used in the main experiments is replaced with a ReEvo-style mechanism to instantiate ASRO’s program-space best-response oracle.
Table~\ref{tab:tsplib_reevo} reports the corresponding ablation results.
\begin{table}[t]
\centering
\caption{\textbf{ReEvo} as the program-space best-response oracle.
Comparison with ReEvo under the same ASRO framework.}
\renewcommand{\arraystretch}{1}
\resizebox{.48\textwidth}{!}{%
\begin{tabular}{@{\hspace{1mm}}lccccc@{\hspace{1mm}}}
\toprule
Method & LIB-S & LIB-M & LIB-L & LIB-XL & Uniform \\
\midrule
ReEvo
& 0.38 & 1.35 & 2.25 & 3.65 & 0.38 \\
ASRO-ReEvo
& \textbf{0.30} & \textbf{1.19} & \textbf{2.06} & \textbf{3.53} & \textbf{0.07} \\
\bottomrule
\end{tabular}%
}
\label{tab:tsplib_reevo}
\end{table}

\subsection{Computation Cost Details}
\label{app:cost}

This section provides additional details for the equal-time comparison. Both ASRO and the EoH baseline are evaluated under an identical wall-clock time budget of 2200 seconds using the same batch evaluation engine described in Appendix~\ref{app:batch_eval}.
Table~\ref{tab:time_budget} reports the resulting optimality gaps under this evaluation protocol.

\begin{table}[t]
\setlength{\tabcolsep}{3pt}
\centering
\caption{
\textbf{Equal-time comparison between EoH and ASRO-EoH on TSP.}
Both methods are evaluated under the same wall-clock time budget (2200s).
}

\label{tab:time_budget}
\begin{tabular}{lccccc}
\toprule
Method & LIB-S & LIB-M & LIB-L & LIB-XL & Uniform \\
\midrule
EoH 
& 0.769 & 1.985 & 3.399 & 4.201 & 0.270 \\
ASRO-EoH 
& \textbf{0.319} & \textbf{0.936} & \textbf{2.417} & \textbf{3.993} & \textbf{0.117} \\
\bottomrule
\end{tabular}
\end{table}

\section{Additional Ablation Studies}

\begin{figure}[!t]
  \centering
  \begin{subfigure}[t]{0.495\linewidth}
    \centering
    \includegraphics[width=\linewidth]{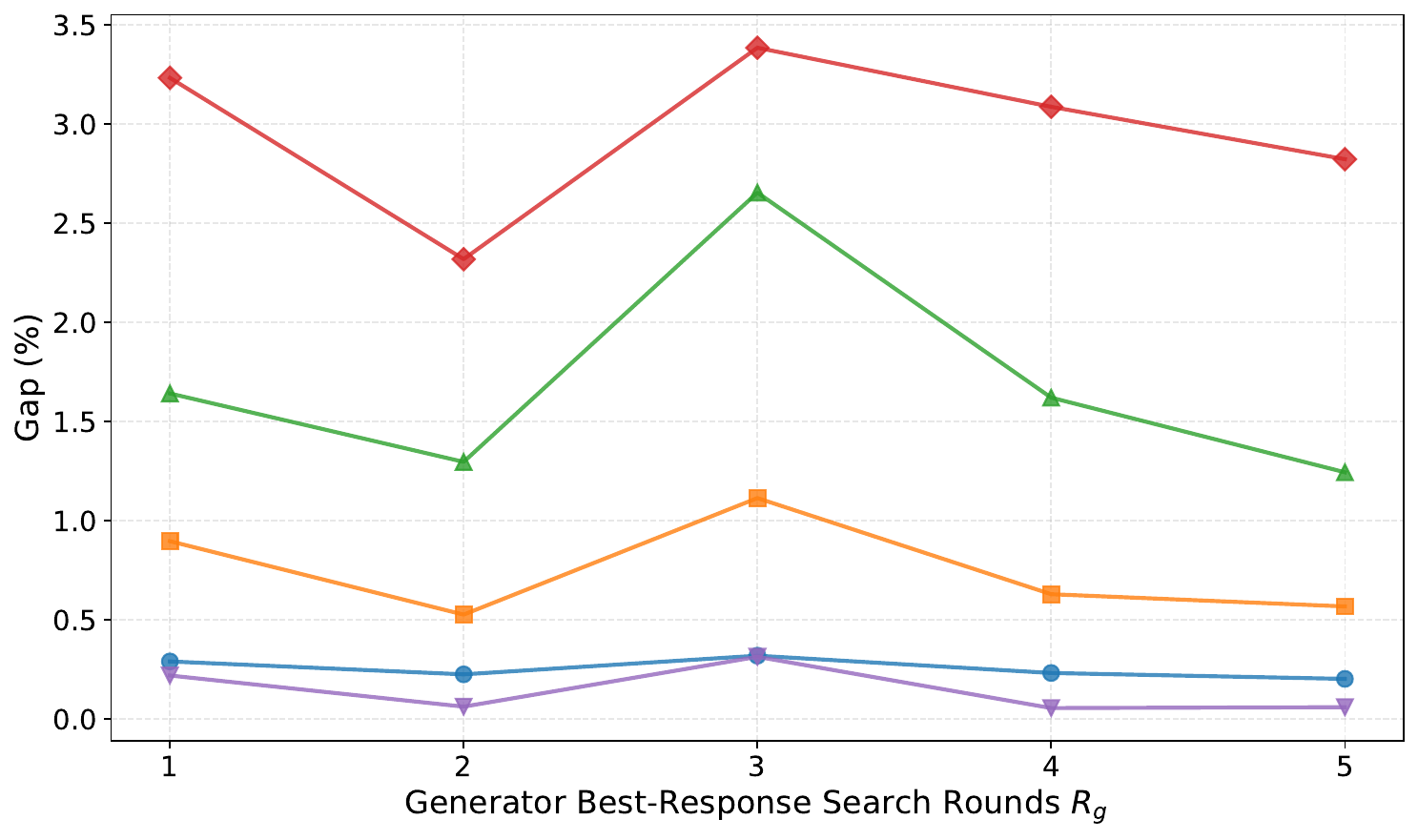}
    \caption{Varying generator best-response rounds $R_g$ with solver rounds fixed ($R_s=2$).}
    \label{fig:ablation_tsp_rg}
  \end{subfigure}\hfill
  \begin{subfigure}[t]{0.495\linewidth}
    \centering
    \includegraphics[width=\linewidth]{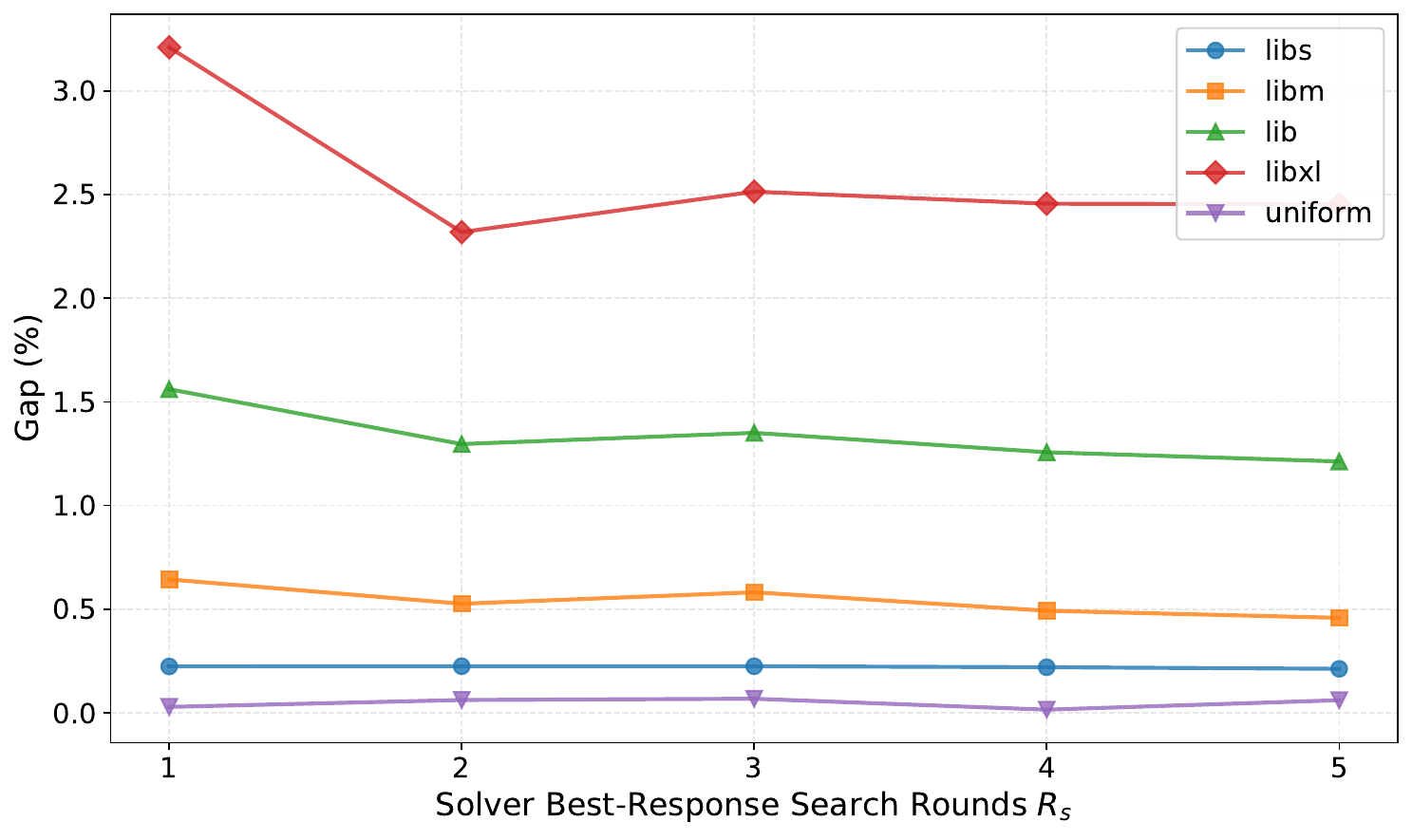}
    \caption{Varying solver best-response rounds $R_s$ with generator rounds fixed ($R_g=2$).}
    \label{fig:ablation_tsp_rs}
  \end{subfigure}

  \caption{\textbf{Ablation of best-response search rounds in ASRO on Euclidean TSP.}
  Each subplot reports the optimality gap (\%) across TSP benchmark instances under different
  configurations of solver-side and generator-side best-response rounds.}
  \label{fig:ablation_tsp_rounds}
\end{figure}

\paragraph{Ablation on Best-Response Search Depth}
We observe heterogeneous sensitivity across benchmarks.
Simpler instance distributions (e.g., uniform instances) are largely insensitive to best-response depth,
while more structured benchmarks (e.g., TSPLIB-XL) exhibit clearer responses.
Increasing solver- or generator-side rounds generally improves best-response accuracy, but with diminishing returns.
On the solver side, performance largely saturates at $R_s=2$.
On the generator side, moderate depth is sufficient; isolated fluctuations may arise due to stochastic effects and conditioning on a fixed counterpart during ablation.
Overall, we adopt $R_s=2, R_g=2$ as a balanced setting that achieves strong performance with substantially lower computational cost.

\begin{figure}[!t]
    \centering
    \includegraphics[width=0.5\linewidth]{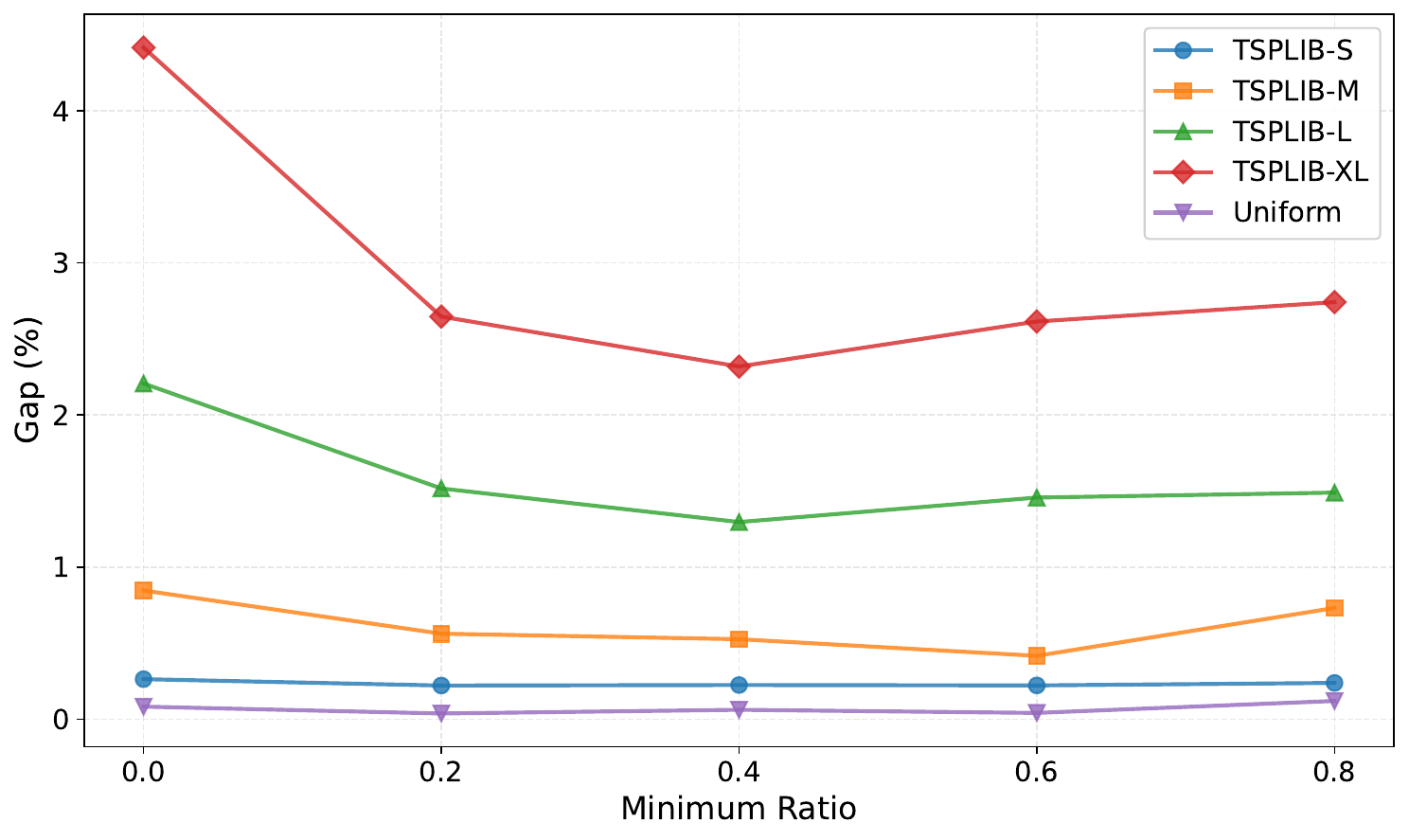}
    \caption{Ablation of the minimum base-generator ratio in ASRO on Euclidean TSP.
    Results report the average optimality gap (\%) across TSP benchmarks under different minimum ratio settings.}
    \label{fig:tsp_min_ratio_ablation}
\end{figure}

\paragraph{Ablation on Base-Generator Mixing Ratio.}
The minimum ratio specifies a lower bound on the probability mass assigned to the base generator, with the remaining mass allocated to the learned generator mixture during solver-side best-response search.
Its effect varies with instance difficulty.
For simpler distributions, performance is largely insensitive to this parameter, as the solver can reliably optimize against the base generator.
In contrast, more structured benchmarks exhibit a clearer dependence: very small ratios may lead to instability due to overfitting to the learned generator mixture, while excessively large ratios bias the opponent distribution toward the base generator and limit the effectiveness of co-evolution.


\end{document}